\documentclass[10pt,twocolumn,letterpaper]{article}

\usepackage[pagenumbers]{cvpr} 

\def\codeurl{https://github.com/beckerio/scahi} 
\def\codeurldisp{github.com/beckerio/scahi}

\usepackage{amsmath,amssymb}
\usepackage{booktabs,caption,subcaption,float,stfloats}
\usepackage{xcolor}
\usepackage{enumitem}
\usepackage{xurl}
\usepackage{tikz}
\usetikzlibrary{arrows.meta,calc,positioning,decorations.pathreplacing,patterns}
\graphicspath{{figs/}}
\setlist{nosep,leftmargin=*}
\definecolor{ink}{HTML}{0B0B0B}
\definecolor{ink2}{HTML}{52514E}
\definecolor{muted}{HTML}{8A8984}
\definecolor{gridc}{HTML}{E3E2DC}
\definecolor{pblue}{HTML}{2A78D6}
\definecolor{pbluel}{HTML}{CDE2FB}
\definecolor{porange}{HTML}{EB6834}
\definecolor{porangel}{HTML}{FBE1D6}
\definecolor{paqua}{HTML}{1BAF7A}
\newcommand{\eps}{\varepsilon}

\usepackage{pgfplots}
\pgfplotsset{compat=1.17}
\newif\ifdraftmarks
\draftmarksfalse 

\usepackage{placeins}

\ifdefined\codeurl\else\def\codeurl{}\fi
\newcommand{\codelocation}{\ifdefined\scahireview The supplementary archive (directory \texttt{code/})\else\ifx\codeurl\empty The code repository\else Our \href{\codeurl}{code repository}\fi\fi}
\newcommand{\codelicencenote}{\ifdefined\scahireview\else\ifx\codeurl\empty\else{} (the SCAHI fusion itself under Apache-2.0)\fi\fi}

\newcommand{\abstractcode}{\ifdefined\scahireview\else\ifx\codeurl\empty\else{} Source code is available in our \href{\codeurl}{code repository}.\fi\fi}
\newcommand{\codesentence}{\ifdefined\scahireview{} Code is included in the supplementary material and will be made publicly
available.\else\ifx\codeurl\empty{} Code will be made publicly available.\else{} Code, frozen configurations and the decision
log are available in our \href{\codeurl}{code repository}; the SCAHI fusion is released under the Apache-2.0 licence.\fi\fi}

\definecolor{cvprblue}{rgb}{0.21,0.49,0.74}
\usepackage[pagebackref,breaklinks,colorlinks,allcolors=cvprblue]{hyperref}
\usepackage{orcidlink}
\usepackage{algorithm}
\usepackage[noend]{algpseudocode}

\def\confName{CVPR}
\def\confYear{2027}

\title{When Does Sliced Inference Pay Off?\\Resolution, Tiling and Dense Fusion for Small Object Detection}

\author{Stefan Becker\,\orcidlink{0000-0001-7367-2519} \qquad Jens Bayer\,\orcidlink{0000-0002-2806-6920} \qquad
David M{\"u}nch\,\orcidlink{0000-0002-8577-5256} \qquad Wolfgang H{\"u}bner\,\orcidlink{0000-0001-5634-6324}\\
Fraunhofer IOSB, Ettlingen, Germany \quad Fraunhofer Center for Machine Learning\\
{\tt\small \{firstname.lastname\}@iosb.fraunhofer.de}
}

\begin{document}
\maketitle
\begin{abstract}
Sliced inference such as SAHI is the standard remedy for small objects in high-resolution imagery, but its gains are
usually measured against low-resolution baselines. We revisit sliced inference with scale as the explicit variable and ask when
tiling actually pays off. We compare nine dense detectors, eight YOLO models of four generations and FCOS, on VisDrone and DOTA, under
pre-specified hypotheses, calibration-only tuning and a single evaluation on held-out test data. Two findings emerge. First,
much of the benefit attributed to slicing is a resolution effect: on VisDrone, a single pass trained at a higher resolution
outperforms tiling with box fusion (merging decoded boxes) for every detector that could be trained at that resolution, and at
equal compute a higher resolution helps
more than a larger model. Second, tiles carry complementary information that box fusion wastes. We introduce SCAHI (Slicing
with Coverage-Aware Hyper Inference), which accumulates the dense outputs of all passes with weights that give every image
position the same total coverage, and reads objects off a peak grid fine enough to keep crowded neighbours apart. On identical
passes, SCAHI outperforms box fusion for every YOLO detector on both datasets and matches it for FCOS. On DOTA, whose images
far exceed the resolution a single pass can be trained at, tiles with SCAHI beat the best single pass for every detector, by
4.3 to 5.6\,mAP50 at 2048\,px, and replacing only SAHI's merge step by SCAHI adds 2.3 to 4.0\,mAP50. An ablation attributes
the gain to the border weighting and the resolution of the peak grid.\abstractcode
\end{abstract}

\section{Introduction}

Tiling a large image into overlapping slices and running the detector on each slice is routine for small objects in drone
and satellite imagery~\cite{akyon2022sahi}. The reported gains are large, but they are usually measured against a full-image
baseline at a low input resolution. This mixes two effects. A tile shows a part of the image at a larger scale, which helps
small objects, but the same effect can be obtained by feeding the whole image at a higher resolution. On
VisDrone~\cite{zhu2022visdrone}, a typical image has a long side of about 1360\,px, and a 640\,px tile at native scale shows
objects at almost the same size as a full image resized to 1280\,px.

We therefore treat scale as the explicit variable and compare every sliced configuration with a single pass trained and
run at a matched or higher resolution. All comparisons use one evaluation code path per dataset, a calibration split for
every tunable choice, hypotheses fixed before the corresponding results existed, and a single evaluation on held-out test
data. We use two datasets that represent two regimes. On VisDrone, images are moderately large, and a single pass at the
native resolution can be trained. On DOTA~\cite{xia2018dota}, images are several thousand pixels wide and densely populated,
and training a single pass beyond 2048\,px exceeds the memory of our GPU.

The answer to the title question depends on this regime. Our study makes two main contributions:
\begin{enumerate}
  \item \textbf{An insight: resolution beats tiling with box fusion as long as the resolution can be trained.} In our experiments on VisDrone,
        a single pass trained at a higher resolution outperforms sliced inference with box fusion for every detector that could be
        trained at that resolution, and at equal
        compute a higher resolution helps more than a larger model.
  \item \textbf{A method: SCAHI.} Tiles nevertheless carry complementary information, but box fusion wastes it. SCAHI, a
        dense fusion of detector outputs with coverage-aware weights and a peak grid that keeps crowded neighbours apart,
        outperforms box fusion on identical passes for every YOLO detector on both datasets and matches it for FCOS. With it,
        tiles beat the strongest trainable single pass by a wide margin on DOTA; on VisDrone they do so only with a finer peak
        grid, found post hoc and confirmed on test-dev.
\end{enumerate}
A component ablation, a decomposition of the gain and scale analyses explain where the gains come from.\codesentence

\begin{figure*}[t]
  \centering
  \includegraphics[width=\textwidth]{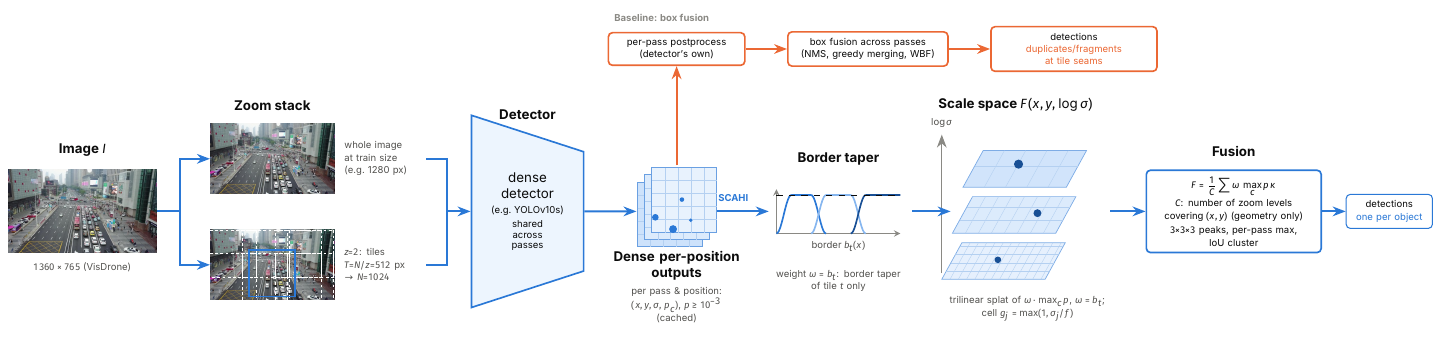}
  \caption{\textbf{Overview.} One detector processes a whole-image pass at its training size and overlapping tiles. The
  baseline merges decoded boxes across passes (orange). SCAHI accumulates the dense outputs of all passes in a global volume
  over position and object size, weights them so that every position receives the same total weight, and extracts one peak
  per object. For YOLOv10 and YOLO26, SCAHI uses the one-to-many head.}
  \label{fig:method}
\end{figure*}

\section{Related work}

\textbf{Sliced inference.} SAHI~\cite{akyon2022sahi,akyon2021sahisoftware} popularised slicing-aided inference and
fine-tuning and is widely applied, e.g.\ to fisheye road cameras~\cite{gia2024fisheye}. Follow-ups make slicing cheaper or adaptive: ASAHI~\cite{moretti2026asahi} fixes the number of slices,
DAHI~\cite{suarezramirez2026dahi} and GOIS~\cite{muzammul2025gois} select dense regions, ROI-Gated
SAHI~\cite{riyadh2026roigatedsahi} selectively applies sliced inference to regions identified by a lightweight object proposer,
skipping background regions when foreground coverage is low, and ESOD~\cite{liu2025esod} slices at feature level. Earlier, ClusDet~\cite{yang2019clusdet}, DMNet~\cite{li2020dmnet} and UFPMP-Det~\cite{huang2022ufpmp} chose crops from
clusters, density maps or a coarse detector. These works compare against full-image baselines at a fixed, usually low, input size; matched-resolution baselines
are rarely reported. Contemporaneous work~\cite{bui2027resolution} reaches a similar conclusion for person detection on
VisDrone with YOLOv8, running weights trained at 640\,px at inference resolutions up to 1536\,px. We train every detector at
the resolution at which it is evaluated, cover several detector families and a second regime, DOTA, in which tiling does pay
off, and propose a fusion that turns the tiles into accuracy.

\textbf{Scale normalisation.} SNIP~\cite{singh2018snip} and SNIPER~\cite{singh2018sniper} train only on objects within a
valid size range per pyramid scale; AutoFocus~\cite{najibi2019autofocus} restricts high-resolution processing to likely
regions, and content-adaptive downsampling~\cite{hesse2023cad} varies the resolution inside the network.

\textbf{Merging detections.} Multi-pass inference needs a merge step: NMS, Soft-NMS~\cite{bodla2017softnms}, SAHI's greedy
non-maximum merging~\cite{akyon2022sahi}, or Weighted Boxes Fusion~\cite{solovyev2021wbf}. All operate on decoded boxes, as does the aggregation of
test-time augmentations~\cite{shanmugam2021tta}, which combines the outputs of several passes after they are decoded.
Voting-based detectors such as HoughNet~\cite{samet2020houghnet} accumulate evidence for objects within a single pass;
SCAHI instead merges the outputs of overlapping passes with coverage-aware weights. Weighted aggregation of overlapping
windows is standard in semantic segmentation, for example Gaussian-weighted sliding
windows in nnU-Net~\cite{isensee2021nnunet}. There, the aggregated map is the output. In detection, individual objects must
afterwards be separated from the aggregated evidence, and, as we show, this step decides whether multi-pass inference pays off.

\textbf{Detectors and data.} YOLOv10~\cite{wang2024yolov10} trains a one-to-many and an NMS-free one-to-one head and uses only
the latter at inference. YOLOv11~\cite{ultralyticssoftware} and YOLOv12~\cite{tian2025yolov12}\footnote{Ultralytics names these models YOLO11 and YOLO12.} use a single dense head with NMS.
All three predict box distributions per position as in generalised focal loss~\cite{li2020gfl}; YOLO26~\cite{jocher2026yolo26}, like YOLOv10, has a one-to-many and an NMS-free one-to-one branch and regresses boxes directly. DOTA~\cite{xia2018dota,
ding2022dota} is the standard benchmark for object detection in large aerial images.

\section{Study design}

\textbf{Datasets and metrics.} On VisDrone2019-DET~\cite{zhu2022visdrone} we use the standard 10 classes and the toolkit
AP@[.5:.95] with ignore regions treated as in the official toolkit, whose evaluation we ported and validated against the
original. On DOTA v1.5~\cite{ding2022dota} we use horizontal boxes, all 16 classes and the official mAP50 of the DOTA devkit,
again through a validated port, without a cap on detections per image. The devkit ignores objects marked \emph{difficult},
which include all objects below 8\,px; a pre-specified secondary evaluation counts them as normal objects. To bound memory,
the DOTA pipeline keeps at most 200{,}000 detections per image (for box fusion, per pass); on VisDrone, every method is capped at
500 detections per image, as in the toolkit. We call an object \emph{crowded} if another object of the same class overlaps it
with IoU $>0.3$, and \emph{isolated} otherwise; the recall of both groups is measured at IoU 0.5.

\textbf{Splits.} On VisDrone, a calibration split of 594 images from 30 whole video sequences is held out of the training set;
we report on val and evaluated test-dev once. On DOTA, a calibration split of 141 training images is held out, and the
official val split (458 images) serves as the held-out test set; its test labels are not public. All model selection and
tuning use the calibration splits.

\textbf{Hypotheses.} Before the respective results existed, we fixed three questions for both datasets, the success threshold
$\eps=0.30$ (AP or mAP50 points) and the selection rules:
\textbf{(Q1)} does a single pass at higher resolution beat tiling with box fusion;
\textbf{(Q2)} does SCAHI beat box fusion on identical passes;
\textbf{(Q3)} do tiles with SCAHI beat the strongest single pass.
Differences $\Delta$ are reported with paired bootstrap 95\,\% confidence intervals over images (1000 resamples, percentile
intervals). On VisDrone, a question holds for a detector if $\Delta>\eps$ (for Q3 also with a lower confidence bound above
zero), and it is confirmed across detectors if it holds for at least four of the five detectors of the extension. On DOTA, a
detector meets a question if $\Delta>\eps$ and a one-sided bootstrap test rejects $\Delta\le\eps$ at $\alpha=0.05$ after Holm
correction over YOLOv10s and YOLOv10m, which must both meet it; Q3 also requires a lower confidence bound above zero. Two further YOLO detectors and, last, FCOS were added to DOTA later under
the same rules; FCOS, a single model, is judged without correction. For the added detectors we report verdicts per detector and
a count over all five. YOLO26s and YOLO26m were added to VisDrone after the verdicts of the other detectors, under a dated
addendum with unchanged questions and rules; their verdicts are reported per detector and do not enter the count. On DOTA, a
fourth question was also fixed before its results
existed: \textbf{(Q4)} does SCAHI improve SAHI when it replaces only SAHI's merge step, on exactly SAHI's passes? Untestable cases count as not confirmed, and post-hoc analyses never enter a verdict.

\textbf{Detectors and training.} VisDrone: YOLOv10s/m, YOLOv11s/m, YOLOv12s/m and YOLO26s/m; DOTA: YOLOv10s/m, YOLOv11s, YOLOv12s and
FCOS~\cite{tian2019fcos} (torchvision implementation, ResNet-50-FPN). All YOLO detectors start from
COCO weights and follow one recipe per dataset (Ultralytics~\cite{ultralyticssoftware}; 100 epochs on VisDrone, 60 on DOTA;
model selection on the calibration split). FCOS also starts from COCO weights but uses a simpler recipe (horizontal flips
only), so absolute accuracies are not comparable across families; all questions compare configurations of the same detector.
Deviations, such as smaller batches for two YOLOv12 runs, are listed in the supplement.

\section{SCAHI}
\begin{figure*}[t]
\centering
\begin{tikzpicture}[x=0.78cm,y=0.9cm,font=\scriptsize,>={Stealth[length=1.8mm]}]
  \foreach \px/\title in {0/{(a) Box fusion (NMS)}, 8.6/{(b) Dense scale-space fusion}}{
    \begin{scope}[xshift=\px cm]
      \node[font=\small\bfseries,anchor=west] at (0,3.35) {\title};
      \fill[gridc!60] (3.5,0) rectangle (5,2.6);
      \draw[ink2,line width=0.6pt] (0,0) rectangle (5,2.6);
      \draw[ink2,line width=0.6pt,densely dashed] (3.5,0.05) rectangle (8.5,2.55);
      \node[ink2,anchor=north west] at (0,2.6) {tile A};
      \node[ink2,anchor=north east] at (8.5,2.55) {tile B};
      \node[muted,font=\tiny] at (4.25,0.2) {overlap};
      \fill[ink!18] (4.3,1.0) rectangle (6.0,1.75);
      \node[ink2,font=\tiny] at (5.15,1.38) {object};
    \end{scope}
  }
  \begin{scope}[xshift=0cm]
    \draw[porange,line width=1.0pt] (4.3,0.95) rectangle (5.0,1.8);
    \draw[porange,line width=1.0pt] (4.25,0.9) rectangle (6.05,1.85);
    \node[porange,anchor=east,align=right] at (3.3,2.05) {A: fragment\\cut at its border};
    \draw[porange,->] (3.35,2.0) -- (4.35,1.75);
    \node[porange,anchor=north,align=center] at (6.9,0.85) {B: full box};
    \node[align=left,anchor=north west,ink] at (0,-0.25) {IoU(fragment, full) $=0.41<0.5$: both survive NMS\\
      $\Rightarrow$ one true positive \textbf{and one false positive} at the seam};
  \end{scope}
  \begin{scope}[xshift=8.6cm]
    \draw[pblue,line width=1.2pt] (4.25,0.9) rectangle (6.05,1.85);
    \node[pblue,anchor=north,align=center] at (6.9,0.85) {one peak, one box};
    \begin{scope}[yshift=-1.55cm]
      \draw[->,ink2] (0,0) -- (8.8,0) node[right]{$x$};
      \draw[ink2] (0,0) -- (0,1.05);
      \node[ink2,anchor=east] at (0,1) {1};
      \node[ink2,anchor=east] at (0,0) {0};
      \draw[pblue,line width=0.8pt,domain=3.5:5,samples=30] (0,1) -- (3.5,1) plot (\x,{0.5*(1+cos(180*(\x-3.5)/1.5))}) -- (5,0);
      \draw[paqua,line width=0.8pt,domain=3.5:5,samples=30] (3.5,0) plot (\x,{0.5*(1-cos(180*(\x-3.5)/1.5))}) -- (8.5,1);
      \draw[ink,densely dashed] (0,1.08) -- (8.5,1.08) node[right,font=\tiny]{$\sum_t b_t=1$};
      \node[pblue,anchor=south] at (1.6,1.0) {$b_A$};
      \node[paqua,anchor=south] at (7.2,1.0) {$b_B$};
      \draw[porange,densely dotted] (4.65,0) -- (4.65,1.25);
      \fill[porange] (4.65,0.21) circle (1.6pt);
      \node[porange,anchor=west,align=left,font=\tiny] at (4.75,0.45) {fragment of A\\weighted $b_A\approx0.2$};
    \end{scope}
    \node[align=left,anchor=north west,ink] at (0,-1.75) {$S_c=\frac{1}{C}\sum_k \max_{i\in\mathcal K\cap k} \omega_i\,p_{c,i}$ (Eq.~\ref{eq:peak}): the partial view of A\\
      near its border is down-weighted, B dominates,\\and the cluster yields a single detection.};
  \end{scope}
\end{tikzpicture}
\caption{\textbf{Why dense fusion removes tile-border fragments.} (a) Box fusion sees two decoded boxes for one object that
straddles a tile seam: the fragment predicted by tile A overlaps the full box from tile B too little to be suppressed. (b) Dense
fusion weights every prediction by a raised-cosine taper $b_t$ over the overlap; the tapers of overlapping tiles sum to one
(Eq.~\ref{eq:taper}), so a prediction near a tile border carries little weight and the cluster around the object yields one detection.
The fusion must also keep adjacent objects apart: its peak search has to be fine relative to object size (\cref{sec:when}). Schematic, not to scale.}
\label{fig:fusion}
\end{figure*}
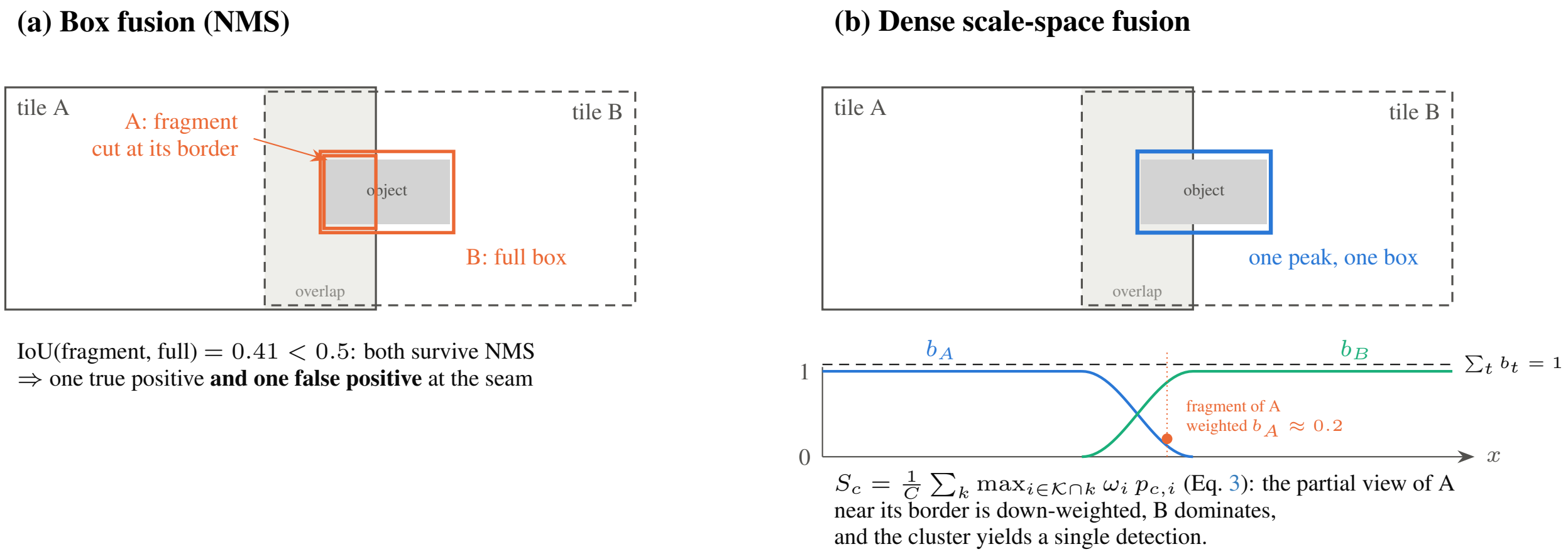

\textbf{Passes.} A configuration consists of passes and a merge rule. A pass sees the image at an absolute zoom $z$, the
number of network-input pixels per image pixel; a tile of side $T=N/z$ image pixels is resized to the network input $N$. Our
multi-pass configuration runs the single pass exactly as trained and adds overlapping tiles with overlap $\rho=0.3$: $z=2$,
$N=1024$ on VisDrone (512\,px tiles) and $z=1$, $N=1024$ on DOTA. The baseline merge is NMS with intersection over the smaller
box, the best box-level rule on calib; greedy merging and Weighted Boxes Fusion merge distinct neighbours in dense scenes.

\textbf{Dense accumulation.} Box fusion sees each pass only through its final boxes. SCAHI combines the passes before boxes
are decided (\cref{alg:scahi,fig:fusion}). Every position $i$ of a pass predicts class probabilities $\mathbf{p}_i$ and a box
with centre $(x_i,y_i)$, width $w_i$ and height $h_i$ in image coordinates, hence object size $\sigma_i=\sqrt{w_ih_i}$. Predictions with
$\max_c p_{c,i}\ge\tau_\text{splat}$ are accumulated in a field over position and log size,
\begin{equation}
F(\mathbf{v})=\frac{1}{C(x,y)}\sum_{k}\sum_{i\in k} \omega_i\,\max_c p_{c,i}\;\kappa(\mathbf{v}-\mathbf{v}_i),
\label{eq:field}
\end{equation}
with $\mathbf{v}=(x,y,u)$ and $\mathbf{v}_i=(x_i,y_i,u_i)$, $u_i=\beta\log_2(\sigma_i/\sigma_0)$, where $k$ runs over the
passes, $\kappa$ is the trilinear interpolation kernel on a grid whose cell size at object size $\sigma$
is $g(\sigma)=\max(1,\sigma/f)$\,px, $\beta$ is the number of size bins per octave, $\sigma_0=2$\,px, $\omega_i$ is the weight of the
prediction and $C$ the coverage, both defined next. For YOLOv10 and YOLO26 we use the one-to-many head, whose redundant per-position
predictions are what the fusion aggregates.

\textbf{Coverage-aware weights.} A prediction close to a tile border sees a truncated object and is unreliable. A tile $t$
therefore weights its predictions with a raised-cosine taper that rises over a width $\Delta=\rho T$ from each of its borders
that lies inside the image,
\begin{equation}
r_t(x,y)=\prod_{e\in\mathcal{E}_t} s(d_e),\quad s(d)=\sin^2\!\bigl(\tfrac{\pi}{2}\,\mathrm{clip}(d/\Delta,0,1)\bigr),
\label{eq:taper}
\end{equation}
where $\mathcal{E}_t$ are the inner borders of $t$ and $d_e$ is the distance of $(x,y)$ to border $e$. The tapers are
normalised over the tiles of a level, $b_t=r_t/\sum_{t'}r_{t'}$, so that $\sum_t b_t(x,y)=1$ at every position: the tiles form a
partition of unity, and the weight of a tile prediction is $\omega_i=b_t(x_i,y_i)$. The whole-image pass has weight $\omega_i=1$, and the
coverage $C(x,y)$ counts the levels of passes that contain $(x,y)$. Every position thus receives the same total weight from each
level, whatever the number of tiles that cover it, and fragments of cut objects receive little weight. Because both levels
cover the whole image, $C=2$ everywhere and the normalisation in \cref{eq:field} is a constant in our configurations; it
matters only for passes that cover part of the image.

\textbf{Separating neighbours.} Objects are read off as local maxima of $F$ with $F\ge\tau_\text{peak}$ in their
$3\times3\times3$ neighbourhood. Around a peak, the predictions within 1.5 cells and one size bin whose boxes overlap the
strongest of them with IoU $\ge0.5$ form a cluster $\mathcal{K}$, across all passes. Its score for class $c$ and its box are
\begin{gather}
S_c=\frac{1}{C}\sum_{k}\max_{i\in\mathcal{K}\cap k} \omega_i\,p_{c,i},\label{eq:peak}\\
\hat{\mathbf{q}}=\frac{\sum_{i\in\mathcal{K}}a_i\,\mathbf{q}_i}{\sum_{i\in\mathcal{K}}a_i},\qquad a_i=\omega_i\max_c p_{c,i},\label{eq:box}
\end{gather}
with box parameters $\mathbf{q}_i=(x_i,y_i,\log w_i,\log h_i)$. A pass that covers the peak but has no member contributes zero, which acts as
negative evidence. Up to three classes with $S_c>0$ are kept per peak, and no NMS follows. The grid must be fine relative to
object size: if one cell covers two adjacent objects, they become one detection. On VisDrone, our pre-specified setting used a
cell factor of $f=5$; a post-hoc analysis (\cref{sec:when}) showed that this merges crowded neighbours, and the
\emph{refined grid}, which selects $f\in\{7,\dots,13\}$, $\beta$ and the thresholds on calib, was pre-specified for DOTA. SCAHI
leaves the detector unchanged and replaces only the merge step. This makes two effects visible that box fusion hides: evidence
from overlapping passes can be combined before objects are decided, and the resolution of the peak search relative to object
size determines whether tiles improve on a strong single pass.

\begin{algorithm}[t]
\caption{SCAHI fusion of one image}
\label{alg:scahi}
\small
\begin{algorithmic}[1]
\Require passes $k=1..K$ (whole image and tiles) with dense predictions $(x,y,w,h,\mathbf{p})$; cell factor $f$; size bins per
  octave $\beta$; thresholds $\tau_\text{splat},\tau_\text{peak}$
\For{each pass $k$ and each prediction $i$ with $\max_c p_{c,i}\ge\tau_\text{splat}$}
  \State map the box to image coordinates; $\sigma_i\gets\sqrt{w_i h_i}$
  \State $\omega_i\gets b_k(x_i,y_i)$ \Comment{\cref{eq:taper}; whole image: $1$}
  \State splat $\omega_i\max_c p_{c,i}$ trilinearly at $\mathbf{v}_i$ \Comment{cell size $\max(1,\sigma/f)$}
\EndFor
\State $F\gets$ accumulated field divided by the coverage $C$ \Comment{\cref{eq:field}}
\For{each local maximum of $F$ in its $3\times3\times3$ neighbourhood with $F\ge\tau_\text{peak}$}
  \State cluster $\mathcal{K}$: nearby predictions with IoU $\ge0.5$ to the strongest, over all passes
  \State class scores $S_c$ and box $\hat{\mathbf{q}}$ \Comment{\cref{eq:peak,eq:box}}
  \State keep up to three classes with $S_c>0$
\EndFor
\end{algorithmic}
\end{algorithm}

\section{Results}

\subsection{VisDrone: resolution beats tiling (Q1)}

\begin{table}[t]
\centering\scriptsize
\setlength{\tabcolsep}{1.9pt}
\caption{\textbf{VisDrone: single pass vs.\ tiling} (AP on val). Single passes trained and run at 1280 and 2048\,px; SAHI;
the 1280\,px pass with tiles merged by box fusion (NMS); and the 2048\,px pass with tiles merged by SCAHI with the pre-specified
setting and with the refined grid ($^\dagger$post hoc on VisDrone). $\Delta$Q1 = single pass at 2048\,px minus tiles with box
fusion, on val and on the held-out test-dev. YOLOv12m could not be trained at 2048\,px. $^\ddagger$Added after the
verdicts of the other detectors under a dated addendum with unchanged rules.}
\label{tab:q1}
\begin{tabular}{@{}lrrrrrrrr@{}}
\toprule
 & \multicolumn{2}{c}{single pass} & & 1280\,px & \multicolumn{2}{c}{2048\,px + tiles} & \multicolumn{2}{c}{$\Delta$Q1} \\
\cmidrule(lr){2-3}\cmidrule(lr){5-5}\cmidrule(lr){6-7}\cmidrule(lr){8-9}
Detector & 1280 & 2048 & SAHI & + tiles, box & SCAHI & refined$^\dagger$ & val & test \\
\midrule
YOLOv10s & 33.66 & 38.47 & 31.52 & 35.00 & 39.34 & \textbf{40.72} & $+3.47$ & $+2.75$ \\
YOLOv10m & 36.81 & 41.03 & 34.18 & 37.66 & 41.29 & \textbf{42.67} & $+3.37$ & $+3.05$ \\
YOLOv11s & 32.20 & 36.80 & 30.26 & 34.05 & 38.66 & \textbf{40.03} & $+2.75$ & $+1.83$ \\
YOLOv11m & 35.82 & 40.00 & 33.80 & 37.05 & 41.54 & \textbf{42.86} & $+2.95$ & $+1.90$ \\
YOLOv12s & 32.64 & 37.94 & 30.95 & 34.87 & 39.80 & \textbf{41.18} & $+3.07$ & $+2.03$ \\
YOLOv12m & 35.41 & --    & 33.70 & 36.95 & --    & --             & --      & --      \\
\midrule
YOLO26s$^\ddagger$ & 33.63 & 38.41 & 31.02 & 34.97 & 39.31 & \textbf{40.65} & $+3.44$ & $+2.54$ \\
YOLO26m$^\ddagger$ & 37.41 & 41.25 & 34.56 & 37.89 & 41.53 & \textbf{42.89} & $+3.36$ & $+2.72$ \\
\bottomrule
\end{tabular}
\end{table}

\begin{figure}[t]
  \centering
  \includegraphics[width=\columnwidth]{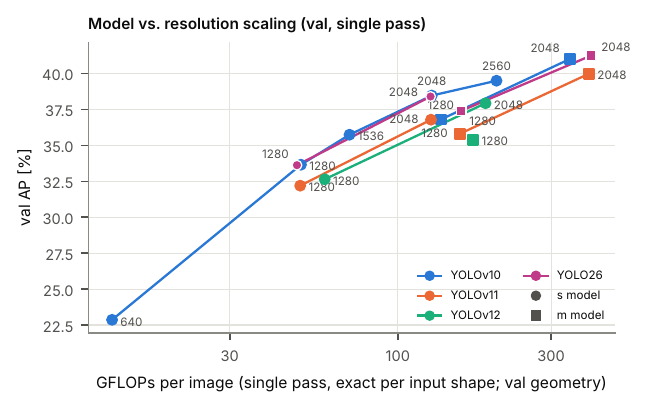}
  \caption{\textbf{Resolution or model size} (VisDrone val). Single-pass AP against compute for each detector at 1280 and
  2048\,px (YOLOv10s also at 640, 1536 and 2560\,px).}
  \label{fig:scaling}
\end{figure}

For every detector that could be trained at 2048\,px (all but YOLOv12m), the single pass beats the 1280\,px pass with tiles
and box fusion, by 2.8--3.5\,AP on val and
1.8--3.1\,AP on test-dev (\cref{tab:q1}); standard SAHI is the weakest configuration. Q1 is confirmed across detectors on both
splits and also holds for the added YOLO26s/m. At almost equal compute, YOLOv10s at 2048\,px (128 GFLOPs) beats YOLOv10m at
1280\,px (136 GFLOPs) by 1.7\,AP on val and 1.6\,AP on test-dev (\cref{fig:scaling}), and YOLO26s at 2048\,px beats YOLO26m at
1280\,px by 1.0 and 0.6\,AP with 20\,\% less compute (126 vs.\ 157 GFLOPs): raising the resolution is the more efficient lever. For YOLOv10s, accuracy still
grows at 2560\,px (39.51\,AP); training at 3072\,px exceeded memory with our fixed batch of three images, and running the
2560\,px weights at 3072\,px adds little (39.64\,AP).

\subsection{VisDrone: fusion and when tiles pay off (Q2, Q3)}
\label{sec:when}

On identical passes, SCAHI beats box fusion for every detector, by 1.4--2.0\,AP on val and 1.9--2.5\,AP on test-dev; Q2 is
confirmed across detectors on both splits. Whether tiles with SCAHI also beat the 2048\,px single pass (Q3) depended on the
detector: with the pre-specified setting it held for three detectors but not for YOLOv10m ($+0.26$\,AP on val, below $\eps$),
and YOLOv12m could not be tested, so Q3 is not confirmed on val. On test-dev, all four testable detectors exceed the threshold;
as verdicts are taken on val, this counts as supporting evidence only. Of the added detectors, YOLO26s meets Q3 (+0.90\,AP), and
YOLO26m, like YOLOv10m the other NMS-free medium model, narrowly misses it (+0.28\,AP).

\begin{figure}[t]
\centering
\begin{tikzpicture}
\begin{axis}[
  width=\columnwidth, height=4.5cm,
  ybar, bar width=3.6pt, ymin=-1.2, ymax=3.9,
  symbolic x coords={YOLOv10s,YOLOv10m,YOLOv11s,YOLOv11m,YOLOv12s,YOLO26s,YOLO26m}, xtick=data,
  x tick label style={font=\tiny, rotate=30, anchor=north east, inner sep=1pt},
  ylabel={$\Delta$AP vs.\ single pass}, ylabel style={font=\scriptsize},
  tick label style={font=\scriptsize},
  legend style={font=\tiny, at={(0.5,1.03)}, anchor=south, legend columns=3, draw=none},
  ymajorgrids, grid style={gridc},
  extra y ticks={0}, extra y tick style={grid style={ink2}},
  enlarge x limits=0.12, area legend,
]
\addplot[fill=pbluel, draw=pblue] coordinates {(YOLOv10s,-0.36) (YOLOv10m,-0.63) (YOLOv11s,1.09) (YOLOv11m,1.10) (YOLOv12s,1.10) (YOLO26s,-0.32) (YOLO26m,-0.57)};
\addplot[fill=pblue, draw=pblue] coordinates {(YOLOv10s,1.05) (YOLOv10m,0.51) (YOLOv11s,2.57) (YOLOv11m,2.29) (YOLOv12s,2.51) (YOLO26s,1.03) (YOLO26m,0.52)};
\addplot[fill=paqua, draw=paqua!70!black] coordinates {(YOLOv10s,1.80) (YOLOv10m,1.25) (YOLOv11s,3.27) (YOLOv11m,2.99) (YOLOv12s,3.32) (YOLO26s,1.78) (YOLO26m,1.29)};
\legend{{SCAHI, single pass only}, SCAHI + tiles, {SCAHI + tiles, refined grid}}
\end{axis}
\end{tikzpicture}
\caption{\textbf{Decomposition on VisDrone test-dev} at 2048\,px (post hoc). Gain over the single pass with standard
post-processing, for SCAHI on the single pass alone, on the single pass plus tiles, and with the refined grid. Tiles add a
similar amount for every detector; fusing the single pass alone helps detectors with NMS and costs the NMS-free YOLOv10 and
YOLO26; the refined grid makes the overall gain consistent.}
\label{fig:decomp}
\end{figure}
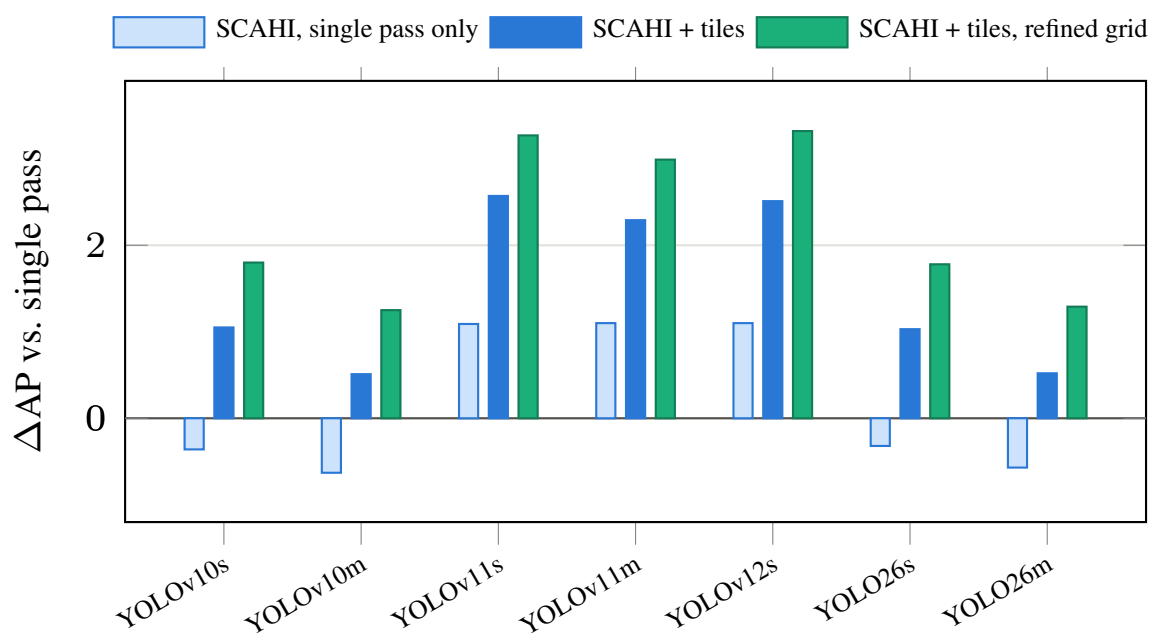

\textbf{Decomposition and over-merging (post hoc).} Splitting the step from the single pass to tiles with SCAHI shows that the
tiles add a similar amount for every detector, 1.1--1.5\,AP at 2048\,px on test-dev (\cref{fig:decomp}), for the original five
almost entirely on objects below 16\,px. The difference lies in fusing the single pass itself: it helps detectors with NMS but
costs the NMS-free YOLOv10 and YOLO26, whose one-to-one heads already separate crowded objects. With a cell factor of~5, the fused single pass loses 20 to
25 points of recall on objects with a same-class neighbour. The refined grid recovers most of it and adds 1.2--1.4\,AP on val
for every detector; on test-dev the gain is about half as large (0.6--0.8\,AP) but positive for every detector, the expected
shrinkage after a search. With it, tiles beat the 2048\,px single pass on test-dev by 1.2--3.3\,AP for every detector.

\textbf{Compute.} For YOLOv10s, a single pass at 2560\,px matches the 2048\,px pass with tiles and the pre-specified SCAHI
on val at 36\,\% of its compute. With the refined grid, tiles exceed it by 1.2\,AP on val and 1.3\,AP [1.05; 1.53] on test-dev,
at 2.8 times the compute (\cref{tab:compute}). On the same tiles, SCAHI costs 77\,ms per image against 51\,ms for box
fusion, mostly because it needs the dense outputs of the detector, and adds 1.9\,AP on val and 2.0\,AP on test-dev with the
refined grid. On VisDrone, tiles therefore buy accuracy, not efficiency.

\begin{table}[t]
\centering\scriptsize
\setlength{\tabcolsep}{2.6pt}
\caption{\textbf{Compute control, YOLOv10s on VisDrone.} AP on val and test-dev, compute per image and latency on one RTX
5090, all rows measured in one exclusive run. Box fusion merges on the CPU, SCAHI on the GPU; SCAHI needs the detector's dense
one-to-many outputs and an unfused model, which is included in its latency; its GPU fusion matches the CPU reference up to
float32 rounding. Rows marked $^\dagger$ are post hoc; the 3072\,px row
is evaluated on val only.}
\label{tab:compute}
\begin{tabular}{@{}lrrrr@{}}
\toprule
Configuration & val & test & GFLOPs & ms \\
\midrule
single pass 2048 & 38.47 & 33.48 & 128 & 9.6 \\
single pass 2560 & 39.51 & 33.98 & 203 & 13.6 \\
\quad run at 3072$^\dagger$ & 39.64 & -- & 298 & 21.0 \\
2048 + tiles, box fusion & 38.85 & 33.25 & 562 & 50.8 \\
2048 + tiles, SCAHI & 39.34 & 34.53 & 562 & 77.3 \\
2048 + tiles, SCAHI refined$^\dagger$ & \textbf{40.72} & \textbf{35.28} & 562 & 77.3 \\
\bottomrule
\end{tabular}
\end{table}

\subsection{DOTA: where tiles pay off}

\begin{table*}[t]
\centering\footnotesize
\setlength{\tabcolsep}{2.6pt}
\caption{\textbf{DOTA v1.5} (mAP50 on the held-out val split, devkit, no cap). Tiles at native resolution, merged by box fusion
or by SCAHI with the refined grid (pre-specified on DOTA). YOLOv12s could not be trained at 2048\,px; its Q3 is measured on its
largest trainable pass ($^\dagger$1024\,px). SAHI + SCAHI: SAHI's own passes (1024\,px weights) merged by SCAHI with the refined
grid instead of SAHI's merge; SAHI rows were not part of the pre-specified FCOS design. Bold: best configuration per detector and input size $N$. Bottom: pre-specified questions,
difference with 95\,\% CI; verdicts with Holm over YOLOv10s/m and, separately, over YOLOv11s/v12s (both pairs also for Q4);
FCOS, a single model, without correction. For YOLOv10, the fusion rows use the one-to-many head; Q1 is pre-specified against box
fusion on the one-to-one head (63.18 and 63.97 at 1024\,px). FCOS uses a simpler training recipe; compare within a column only.}
\label{tab:dota}
\begin{tabular}{@{}llrrrrr@{}}
\toprule
 & Configuration & YOLOv10s & YOLOv10m & YOLOv11s & YOLOv12s & FCOS \\
\midrule
1024\,px & single pass & 55.80 & 58.44 & 56.16 & 56.36 & 42.64 \\
 & + tiles, box fusion & 64.86 & 64.57 & 62.14 & 62.88 & 52.56 \\
 & + tiles, SCAHI & \textbf{66.17} & \textbf{67.34} & \textbf{65.26} & \textbf{65.97} & \textbf{52.96} \\
 & SAHI & 59.20 & 60.96 & 57.90 & 60.21 & -- \\
 & SAHI + SCAHI & 62.08 & 63.22 & 61.87 & 63.30 & -- \\
\midrule
2048\,px & single pass & 65.50 & 67.39 & 65.19 & -- & 55.01 \\
 & + tiles, box fusion & 68.46 & 68.13 & 66.68 & -- & 60.49 \\
 & + tiles, SCAHI & \textbf{71.00} & \textbf{71.73} & \textbf{69.81} & -- & \textbf{60.59} \\
\midrule
Q1 & 2048 vs.\ 1024 + tiles, box & $+2.31$ [$-0.31$; $4.95$] & $+3.42$ [$0.74$; $5.59$] & $+3.05$ [$0.55$; $5.81$] & untestable & $+2.45$ [$-0.49$; $5.28$] \\
Q2 & SCAHI vs.\ box, 1024 & $+1.31$ [$0.36$; $2.39$] & $+2.77$ [$1.67$; $3.83$] & $+3.12$ [$2.02$; $4.18$] & $+3.09$ [$2.11$; $4.43$] & $+0.40$ [$-0.92$; $1.79$] \\
Q2 & SCAHI vs.\ box, 2048 & $+2.54$ [$1.43$; $3.34$] & $+3.60$ [$2.29$; $4.57$] & $+3.12$ [$2.00$; $3.94$] & untestable & $+0.09$ [$-0.91$; $1.37$] \\
Q3 & tiles + SCAHI vs.\ single pass & $+5.50$ [$3.55$; $7.77$] & $+4.34$ [$3.02$; $6.19$] & $+4.62$ [$2.76$; $6.62$] & $+9.62^\dagger$ [$7.10$; $12.22$] & $+5.57$ [$3.77$; $7.65$] \\
Q4 & SCAHI vs.\ SAHI's merge & $+2.87$ [$2.13$; $4.33$] & $+2.26$ [$1.08$; $3.25$] & $+3.97$ [$2.93$; $5.06$] & $+3.10$ [$2.23$; $4.08$] & -- \\
\bottomrule
\end{tabular}
\end{table*}

DOTA represents the other regime (\cref{tab:dota}). Its images are several thousand pixels wide and contain up to ten thousand
objects. Because of this density, training a single pass at 3072\,px exceeded the memory of our GPU even with a batch of two,
and YOLOv12s could not be trained beyond 1024\,px. Here tiles pay off. Because the refined grid was fixed before any DOTA result existed, DOTA tests it on independent data.
Tiles at native resolution with SCAHI beat the strongest
trainable single pass for all five detectors: by 4.3 to 5.6\,mAP50 at 2048\,px, and by 9.6 for YOLOv12s on its weaker 1024\,px
pass (Q3). Counting the difficult objects, the gains grow (e.g. 6.3 and 5.0 for YOLOv10s/m), and AP@[.5:.95] shows the same
picture (supplement). For the YOLO detectors, SCAHI beats box fusion on identical passes at every testable resolution (Q2); for FCOS, both fusions
perform on par ($+0.4$ and $+0.1$\,mAP50, not confirmed). Q1 holds for YOLOv10m and YOLOv11s but narrowly misses for YOLOv10s
and FCOS, whose confidence intervals include zero: on DOTA, higher resolution
alone no longer beats tiling clearly. The verdicts of the original two-detector study are unchanged: Q2 and Q3 confirmed,
Q1 not. Even box fusion with tiles is ahead of the single pass at 2048\,px; SCAHI adds another 2.5 to 3.6\,mAP50.
A post-hoc check indicates that this regime is not explained by inference memory alone: run on the whole image at native
resolution, which fits into memory for 98\,\% of the DOTA-val images (85\,\% of the objects), the 2048\,px weights of
YOLOv10s and YOLOv11s stay 4.1 to 5.9\,mAP50 behind tiles with SCAHI, losing most on objects above 256\,px. Fused by SCAHI
with the 2048\,px pass, the same native pass matches tiles with SCAHI (supplement): what pays off is combining a downscaled view
with native-resolution detail, and tiles provide the latter within memory.

\textbf{SAHI with SCAHI (Q4).} SCAHI also works as a drop-in replacement for SAHI's merge step, given access to the detector's dense outputs: on exactly
SAHI's passes, it adds 2.3 to 4.0\,mAP50 for all
four detectors (Q4, confirmed for both pairs; 3.4 to 4.3 points in AP@[.5:.95]), with large gains on objects above
128\,px, which slicing cuts into fragments, and without adding substantial latency (supplement). For YOLOv10, SAHI uses the
one-to-one and SCAHI the one-to-many head; YOLOv11s and YOLOv12s have a single head, so SAHI and SCAHI see the same predictions.
Their gains (4.0 and 3.1) are at least as large, so they stem from the merge step.

\textbf{Compute on DOTA.} The price of tiles is compute: for YOLOv10s about 5.5 times the GFLOPs and 4.8 times the latency of the single pass (134 vs.\ 28\,ms with the
fusion on the GPU). Of this time, 90\,\% are the tile passes, which every tiled method shares, including SAHI (195\,ms); SCAHI
itself adds about 13\,ms.

\textbf{FCOS.} The non-YOLO detector reproduces the main pattern: tiles with SCAHI beat its single pass by 5.6\,mAP50 and
even its best single pass of the pre-specified inference crossing (2048\,px weights run at 2560\,px) by 2.5\,mAP50. As for
the YOLO detectors, the gain comes from the tiles, not from fusing the single pass. Where SCAHI does not exceed box fusion,
the peak grid explains it (post hoc): on calib, the search retained the coarse pre-specified cell factor for FCOS, and with it
SCAHI recalls about 30 points fewer crowded objects than box fusion, while isolated objects are recalled almost equally well,
the over-merging analysed on VisDrone (Sec.~\ref{sec:when}). Neither the detection cap (capping both fusions equally leaves
the result unchanged) nor a lack of redundant predictions (FCOS yields almost twice as many per object as the YOLO detectors)
accounts for it. The FCOS comparison of the two fusions is also sensitive to the pipeline limit, which caps single very dense
images: without it, a few such images add millions of low-scoring detections that dilute the dataset-wide ranking, SCAHI falls
behind box fusion at 1024\,px ($-1.8$\,mAP50), and the gain of Q3 shrinks to 4.1\,mAP50 but remains clear (supplement).

\subsection{What makes SCAHI work}

\begin{table}[t]
\centering\scriptsize
\setlength{\tabcolsep}{2.5pt}
\caption{\textbf{SCAHI component ablation} (post hoc; 2048\,px pass with tiles). Difference to the pre-specified setting
(border taper, normalisation, cell factor 5) in AP on VisDrone val (range over five detectors) and mAP50 on DOTA-val
(YOLOv10s / YOLOv10m), and change in recall of objects with a same-class neighbour.}
\label{tab:ablation}
\begin{tabular}{@{}lrrr@{}}
\toprule
Variant & VisDrone & DOTA & crowded recall \\
\midrule
without normalisation by $C$ & $\pm0.00$ & $\pm0.00$ / $\pm0.00$ & $\pm0$ \\
without border taper & $-0.74$ to $-1.01$ & $-1.12$ / $-0.99$ & $-2$ to $+1$ \\
with validity windows & $-1.61$ to $-1.79$ & --$^\ast$ & $-1$ to $-2$ \\
refined peak grid & $+1.32$ to $+1.38$ & $+0.36$ / $+0.41$ & $+18$ to $+24$ \\
\bottomrule
\end{tabular}
\\[2pt]{\tiny $^\ast$Windows sized for VisDrone; on DOTA they cut off large objects ($-19$ / $-20$\,mAP50).}
\end{table}

\Cref{tab:ablation} isolates the parts of SCAHI. Removing the explicit normalisation leaves the results unchanged, because
the coverage-aware tapers already give every position the same total weight; removing the tapers themselves costs up to 1.1
points on both datasets. Validity windows that favour object sizes per pass consistently hurt and are not part of SCAHI. The resolution
of the peak grid is the decisive design choice for crowded scenes: a finer grid raises the recall of crowded objects by 18 to
24 points. On VisDrone this lifts AP by 1.3 to 1.4 points; on DOTA, whose primary metric ignores the smallest, most crowded
objects, the gain in mAP50 is smaller, but vehicles, the most crowded classes, gain 3.7 points. Giving box fusion the same
coverage-aware border weights recovers only part of the gap for the YOLO detectors (post hoc, YOLOv10s and YOLOv11s): weighted
NMS gains 0.5 to 0.75\,AP over box fusion on VisDrone and 1.4 to 1.7\,mAP50 on DOTA, but SCAHI remains ahead on the same
passes by 1.0 to 2.3\,AP on VisDrone and 1.1 to 1.4\,mAP50 (2.0 to 2.3\,AP@[.5:.95]) on DOTA. Most of its gain therefore comes
from merging the dense outputs before boxes are decided, which mainly improves localisation. For FCOS, weighting does not help
box fusion, and SCAHI with its coarse grid stays on par with both; weighted Weighted Boxes Fusion falls below plain box fusion
on both datasets (supplement).

\subsection{Scale analyses}

Two analyses on VisDrone clarify what limits small objects. Recall as a function of object size in the network input is
the same whatever the inference resolution, so tiles and a higher input resolution help in the same way; below
about 6 input pixels almost nothing is found. A degradation test suggests that the remaining difficulty of small objects stems
more from the objects themselves, often occluded and crowded, than from missing pixels: large objects that lose half their
resolution lose only 0.03 in recall, while originally small objects at the same apparent size are 0.09 below them. Training
only on objects within a valid size band per zoom level, in the spirit of SNIPER, brought no gain over ignore-aware training.

\section{Discussion and limitations}

\textbf{When does sliced inference pay off?} We observed two regimes. Where the image could be processed at the needed
resolution in a single pass, as on VisDrone, a higher input resolution was the cheaper route and outperformed standard tiling
for every detector that could be trained at it. Where the resolution a single pass could be trained at fell short of the image, as on DOTA, tiles paid off
clearly. A post-hoc check on DOTA refines this: where the native image fits into one inference pass, a native and a downscaled
pass fused by SCAHI match tiles with SCAHI, so what decides is the fusion of zoom levels, and tiles obtain the same native
detail within memory. Which regime applies depends on image size, object density and the available memory, so the boundary will shift
with hardware and detector. For the YOLO detectors, a dense fusion that keeps neighbours apart turned the tiles into more
accuracy than box fusion in both regimes; for FCOS, both fusions were on par, and the reason, a peak grid too coarse for
crowded scenes, is the same mechanism that the refined grid addresses. Per image, the gain of tiles with SCAHI increased with the share of small objects on DOTA (Spearman
$\rho=0.15$ to $0.21$); on VisDrone this relation was weak ($|\rho|\le0.10$), and its relation to the share of
crowded objects was weak at most in both datasets (supplement). Like any tiled inference, including SAHI, SCAHI multiplies the number of passes; most of
its cost comes from these passes, and the fusion adds the remainder. It therefore targets accuracy-oriented processing of large
images, such as offline analysis of aerial imagery; where latency matters more than the last points of accuracy, a single pass
at the highest trainable resolution remains the better choice.

\textbf{Limitations.} (i) SCAHI requires dense per-position predictions, as
produced by anchor-free heads (YOLO, FCOS~\cite{tian2019fcos}, GFL~\cite{li2020gfl}); query-based detectors such as RT-DETR output a fixed set of predictions per
pass and would require a sparse variant of the fusion, which we leave for future work. Apart from FCOS on DOTA, our experiments use YOLO detectors. (ii) On VisDrone, the refined peak grid was found post
hoc; its held-out gain is about half of its val estimate, and on DOTA it was pre-specified. Its cell factor is selected by
mAP50 on a calibration split, which can miss crowded scenes when they are rare, as for FCOS; a selection criterion that
accounts for crowded recall is left for future work. (iii) The DOTA devkit ignores the
smallest objects, and the images we used are re-encoded JPEGs. (iv) All latencies were measured on a single RTX 5090, each table in one exclusive run. (v) Horizontal boxes only.

\section{Conclusion}

In our experiments, whether sliced inference pays off depended on the regime. On VisDrone, where a single pass could be
trained at the needed resolution, it beat tiling with box fusion. On DOTA, where it could not, tiles added substantial accuracy
for every detector; where the native image fitted into one inference pass, a native and a downscaled pass fused by SCAHI
matched them, so what pays off is combining zoom levels, and tiles obtain native detail within memory. For the YOLO detectors,
the merge step decided in both regimes: a dense fusion with coverage-aware weights and a peak grid that keeps neighbours apart
turns this detail into accuracy that box fusion leaves on the table.

\section*{LLM Usage Statement}

The research question and core idea originated with the authors. Under their direction, a general-purpose assistant (Claude,
Anthropic) helped draft experimental designs, analysis plans, manuscript text and two schematic figures and one chart, which the authors
reviewed, revised and approved; Claude Opus 5.5 (Anthropic) was used as a coding agent to implement code and run experiments.
All key decisions, including hypotheses and success criteria, data and models, deviations, frozen configurations and the
single authorized evaluation on held-out test data, were made by the authors and are recorded in a dated decision log.
Experimental results were checked using reproducible scripts and automated tests, including evaluation ports validated
against the official implementations. The authors verified the manuscript and references and take full responsibility for
the work.

{
    \small
    \bibliographystyle{ieeenat_fullname}
    \bibliography{refs}
}

\clearpage
\setcounter{topnumber}{4}\setcounter{dbltopnumber}{4}\setcounter{bottomnumber}{2}\setcounter{totalnumber}{6}
\renewcommand{\topfraction}{0.95}\renewcommand{\dbltopfraction}{0.95}\renewcommand{\bottomfraction}{0.6}
\renewcommand{\textfraction}{0.05}\renewcommand{\floatpagefraction}{0.75}\renewcommand{\dblfloatpagefraction}{0.75}
\raggedbottom
\setcounter{page}{1}
\maketitlesupplementary

\section*{Overview}

This supplement collects material that does not fit into the page limit of the main paper and the information needed to
reproduce the study. \Cref{sec:s-protocol} describes the study protocol and the computing environment. \Cref{sec:s-ext} and
\cref{sec:s-main} give the detailed VisDrone results of the model extension and of the main study with YOLOv10s.
\Cref{sec:s-dota} reports the DOTA study, \cref{sec:s-ablation} the component ablation of SCAHI, \cref{sec:s-qual}
qualitative examples, \cref{sec:s-code} the code and its use, \cref{sec:s-impact} the societal impact, and \cref{sec:s-dev} all deviations from the original plan. Unless stated otherwise, VisDrone numbers
are the toolkit AP@[.5:.95] and DOTA numbers the devkit mAP50, in percent.

\section{Study protocol and reproducibility}
\label{sec:s-protocol}

\textbf{Evaluation.} Our Python port of the official VisDrone toolkit (\texttt{evalDET.m}) reproduces the original
implementation to within $2\cdot10^{-4}$ AP on the full val split, for synthetic and for real detections. Ground truth and
detections that lie mostly in ignore regions are removed as in the toolkit; every method is capped at 500 detections per image,
sorted by score. Size-bin results use pycocotools~\cite{lin2014coco} on the same filtered data. Boxes of the cached whole-image passes
are mapped back to the image with a single scale factor, whereas the Ultralytics predictor uses one per axis and a slightly
different resize rule. We recomputed every frozen row that uses a cached whole-image pass with the Ultralytics conventions,
against a threshold of 0.05 fixed before the check. No number of the main paper and no row that enters a verdict changes by
more than 0.04 (AP@[.5:.95] on VisDrone, mAP50 on DOTA); fifteen exploratory rows of SCAHI on the whole-image pass alone change by
at most 0.10 on the held-out splits. The frozen values are kept; the recomputed values are listed with the code. The YOLO26 rows, added later, use the Ultralytics
conventions from the start.

\textbf{Pre-specification.} Both studies fixed their hypotheses, success thresholds and selection rules in writing before the
corresponding results existed, and recorded every later decision with date and reason in a decision log. Additions after the
start of a study were only allowed to tighten a criterion or to define untestable cases; all of them predate the results they
concern. In the main study, the reference configuration, the candidate set, three hypotheses and the noise threshold were
fixed before the first method training; a second, stronger reference was added before any method result on val existed. In the
extension, the questions Q1--Q3 of the main paper and a descriptive comparison of the two YOLOv10 heads, the counting rule (at least four of five detectors) and the treatment of
untestable cases were fixed before training.

\Cref{tab:s-status} summarises which results are pre-specified and which are post hoc.
\begin{table*}[tp]
\centering\footnotesize
\setlength{\tabcolsep}{4pt}
\caption{\textbf{Status of the reported results.} Pre-specified: questions, rules and settings fixed in writing before the
corresponding results existed; post hoc: defined after the relevant results were known (exploratory, never part of a verdict).}
\label{tab:s-status}
\begin{tabular}{@{}llll@{}}
\toprule
Result & Dataset & Status & Where (paper or this supplement) \\
\midrule
Q1--Q3 verdicts, model extension (five detectors) & VisDrone & pre-specified; val, test-dev once & paper Sec.~5.1--5.2; Sec.~\ref{sec:s-ext} \\
Q1--Q3 verdicts, YOLO26s/m (addendum) & VisDrone & pre-specified; val, test-dev once & paper Sec.~5.1--5.2; Sec.~\ref{sec:s-y26} \\
Criteria of the main study (YOLOv10s) & VisDrone & pre-specified; val, test-dev once & Sec.~\ref{sec:s-main} \\
SCAHI, pre-specified setting ($f=5$, tuned on calib) & both & pre-specified & paper Sec.~4 \\
Refined peak grid ($f=7$--$13$, selected on calib) & VisDrone & post hoc; frozen before test-dev & paper Sec.~5.2 \\
Refined peak grid & DOTA & pre-specified & paper Sec.~5.3 \\
Q1--Q3 verdicts, YOLOv10s/m & DOTA & pre-specified; DOTA-val once & paper Sec.~5.3 \\
YOLOv11s, YOLOv12s, FCOS & DOTA & pre-specified by dated addenda & paper Sec.~5.3 \\
Q4: SCAHI replacing SAHI's merge & DOTA & pre-specified & paper Sec.~5.3 \\
Decomposition, over-merging, per-image analysis & both & post hoc & paper Sec.~5.2, 6 \\
Component ablation, border-weighted box fusion & both & post hoc, after the held-out evaluation & paper Sec.~5.4; Sec.~\ref{sec:s-ablation} \\
Pipeline-limit check & DOTA & check pre-specified, values exploratory & Sec.~\ref{sec:s-dota} \\
FCOS diagnostics & DOTA & post hoc & paper Sec.~5.3; Sec.~\ref{sec:s-dota} \\
Native single pass, multi-zoom without tiles, FCOS control & DOTA & post hoc & paper Sec.~5.3; Sec.~\ref{sec:s-dota} \\
Compute control; 3072\,px row & VisDrone & descriptive; 3072\,px row post hoc (val only) & paper Tab.~2 \\
Latency & both & descriptive, one exclusive run per table & paper Sec.~5.2--5.3; Sec.~\ref{sec:s-dota} \\
Qualitative examples & both & selection rule fixed before rendering & Sec.~\ref{sec:s-qual} \\
\bottomrule
\end{tabular}
\end{table*}

\textbf{Frozen evaluation.} Before each test-dev evaluation, all configurations to be evaluated were frozen together with the
checksums of their weights and the origin of every choice (fixed by design or selected on calib). A replay through the test-dev
code path then had to reproduce the val result of every frozen configuration exactly before test-dev was touched. Test-dev was evaluated once per study and per addendum, and every frozen
configuration is reported.

\textbf{Environment and compute.} Python 3.11, PyTorch 2.14.1 (CUDA 13.0), Ultralytics 8.4.183, SAHI 0.12.8. All runs used one
NVIDIA RTX 5090 (32\,GB) at the default power limit; the study moved once between two identical machines, and the frozen
weights reproduced every val result exactly on both. All latencies were measured on one machine with exclusive use of the GPU, each table in one run. Parts of the calibration tuning were
parallelised; a cross-check confirmed identical selections and identical results. The main study
took about 24 GPU-hours plus CPU time for the fusion.

\FloatBarrier
\section{VisDrone model extension: detailed results}
\label{sec:s-ext}

\begin{figure*}[tp]
  \centering
  \includegraphics[width=\textwidth]{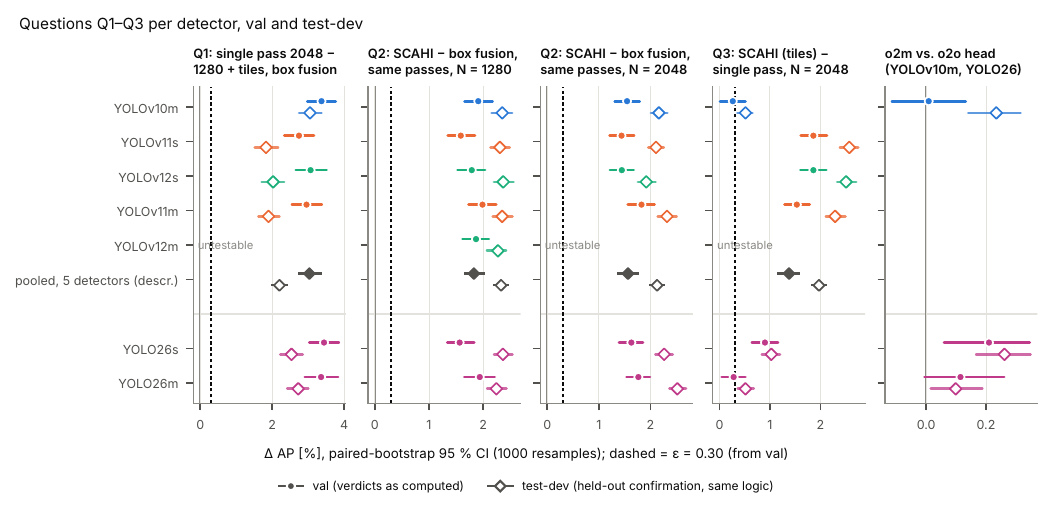}
  \caption{\textbf{Pre-specified questions across detectors} (VisDrone model extension). Differences with paired bootstrap 95\,\% confidence intervals on val (filled) and test-dev (hollow); dashed line: threshold $\eps=0.30$\,AP. Q2 is shown for passes at 1280 and 2048\,px; the last panel compares the two heads under SCAHI for YOLOv10m and YOLO26. The pooled row covers the five detectors of the original extension; YOLO26s/m, added by addendum, are shown below it. YOLOv12m could only be tested at 1280\,px, as it could not be trained at 2048\,px. YOLOv10s, the detector of the main study, is
  reported in \cref{sec:s-main}.}
  \label{fig:forest}
\end{figure*}

\Cref{fig:forest} summarises the pre-specified questions of the extension on val and test-dev.

This section complements Sec.~\ref{sec:when} of the main paper. The decomposition, the over-merging analysis and the refined peak
grid were defined after the val results of the extension were known; they are reported separately from the pre-specified
verdicts and never enter them.

\begin{figure*}[tp]
  \centering
  \includegraphics[width=\textwidth]{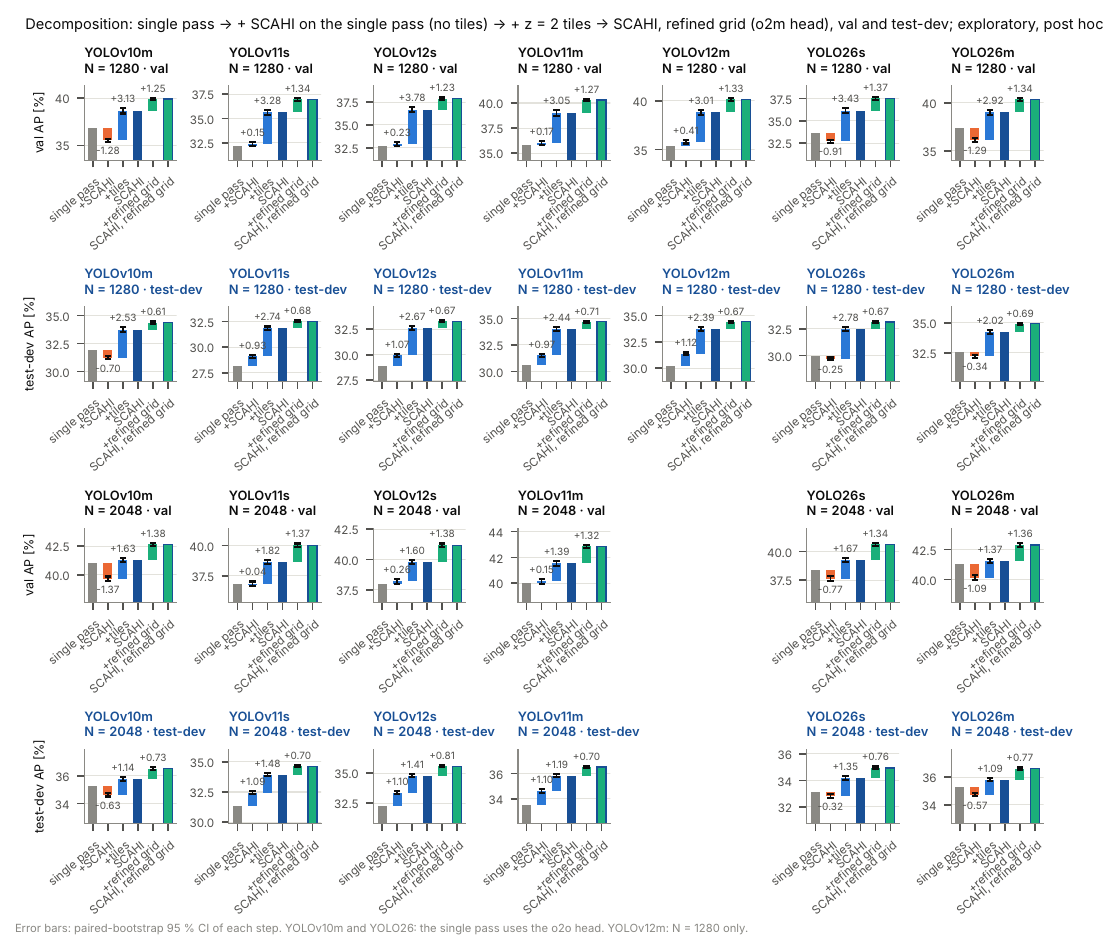}
  \caption{\textbf{Decomposition of the gain of tiles with SCAHI} for the detectors of the extension, including YOLO26, on val and test-dev, at 1280\,px (top) and 2048\,px (bottom): single pass, then SCAHI on the single pass, then adding $z{=}2$ tiles, then the refined grid. Error bars: paired bootstrap 95\,\% confidence intervals of each step. For YOLOv10m and YOLO26, the single pass uses the one-to-one head; YOLOv12m only at 1280\,px, as it could not be trained at
  2048\,px. YOLOv10s is shown in paper Fig.~4.}
  \label{fig:s-waterfall}
\end{figure*}

\textbf{Decomposition.} \Cref{fig:s-waterfall} shows the full decomposition for the detectors of the extension, including YOLO26, both input sizes $N$ and both
splits. Each panel starts from the single pass with its standard post-processing (A), adds SCAHI on the single pass, then the
tiles, and finally the refined peak grid. Error bars are paired bootstrap 95\,\% confidence intervals of each step. The step
from adding tiles is positive in every panel, which is the basis of the statement that tiles carry consistent information for
small objects. The step from fusing the single pass is negative for the NMS-free YOLOv10 and YOLO26, whose one-to-one heads already deliver one
box per object, and positive on test-dev for all detectors with NMS. On test-dev, the refined grid improves on the pre-specified
setting by 0.61--0.81\,AP for all fifteen comparable configurations, about half of its val gain (ratio 0.49--0.59).

\begin{figure*}[tp]
  \centering
  \includegraphics[width=0.85\textwidth]{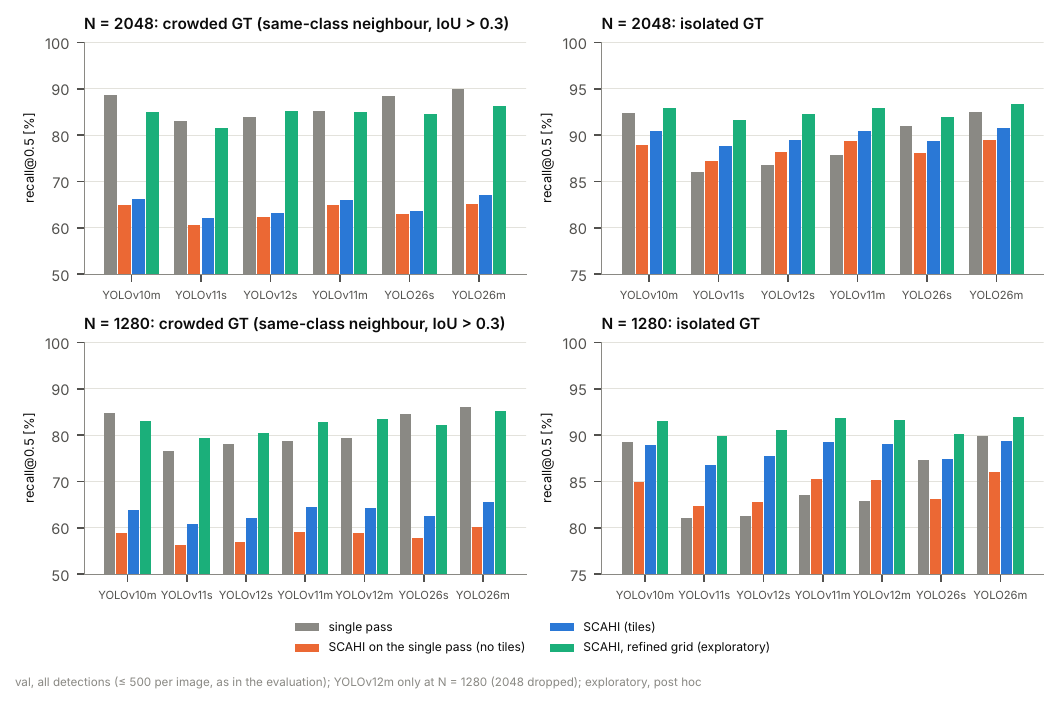}
  \caption{\textbf{Recall of crowded and isolated objects} (IoU 0.5, val) for the single pass (A), SCAHI on the single pass (B), tiles with SCAHI (C) and the refined grid, for the detectors of the extension; YOLOv12m only at 1280\,px, as it could not be trained at 2048\,px.}
  \label{fig:s-recall}
\end{figure*}

\begin{figure*}[tp]
  \centering
  \includegraphics[width=\textwidth,height=0.86\textheight,keepaspectratio]{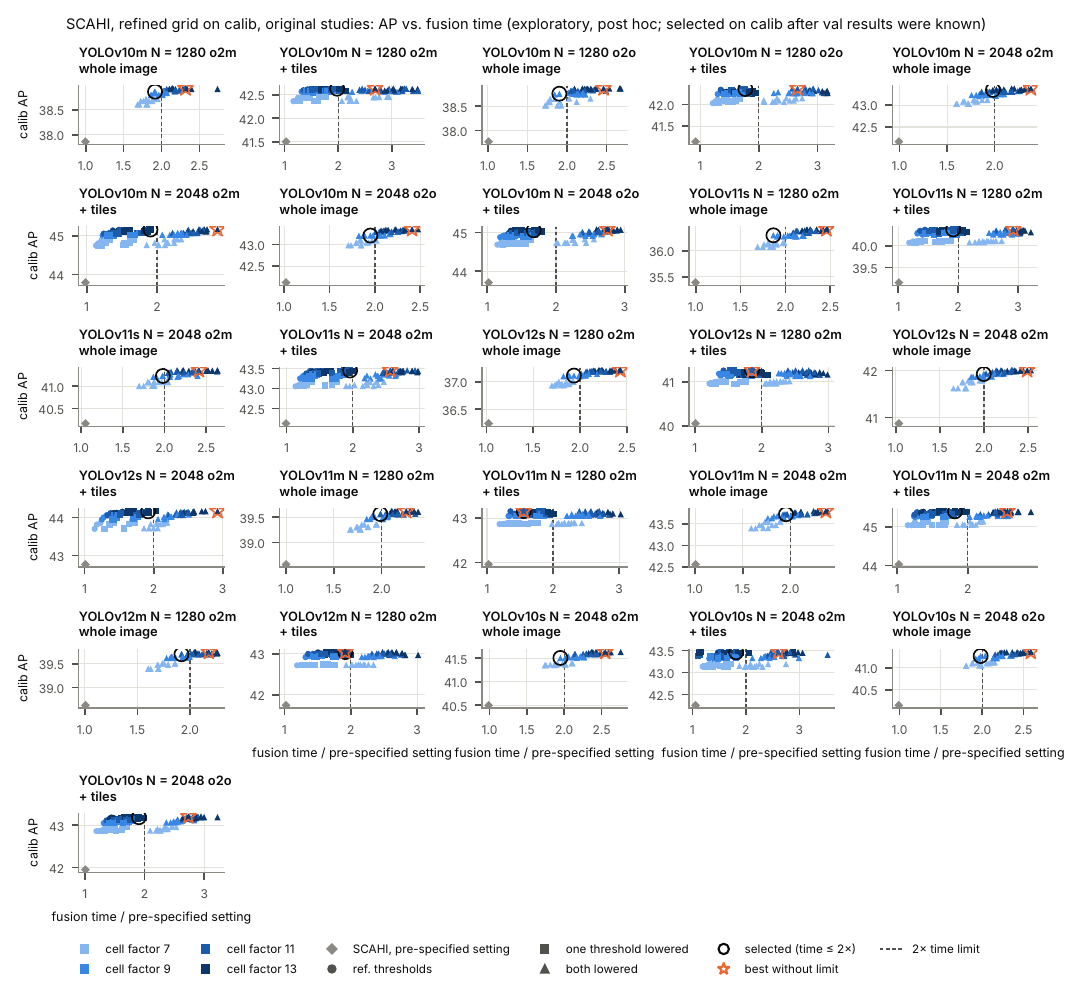}
  \caption{\textbf{Search of the refined peak grid on calib}, detectors of the main study and the original extension. AP against
  fusion time per image for every searched configuration, per detector, input size $N$ and row (B: single pass, C: with tiles);
  the selected configuration is marked. The time limit was twice the fusion time of the pre-specified setting.}
  \label{fig:s-xgrid}
\end{figure*}

\begin{figure*}[tp]
  \centering
  \includegraphics[width=0.8\textwidth,height=0.8\textheight,keepaspectratio]{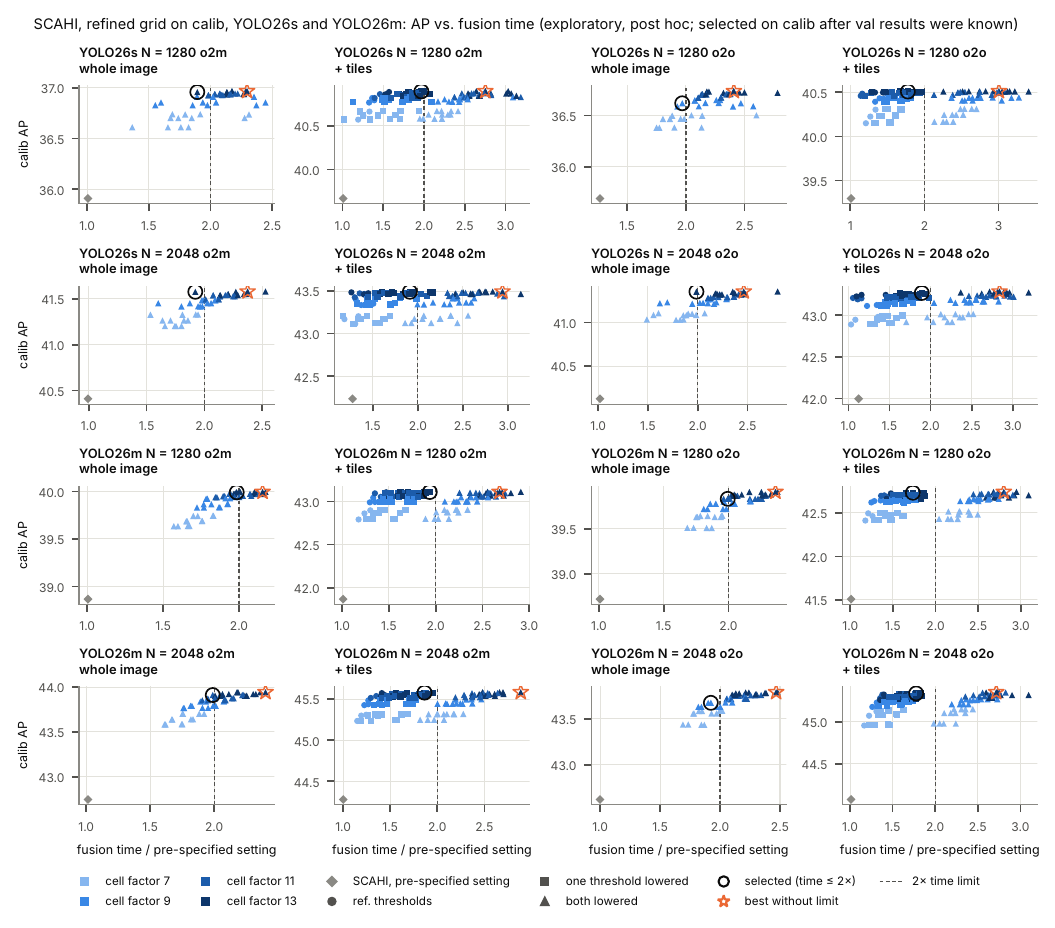}
  \caption{\textbf{Search of the refined peak grid on calib}, YOLO26s and YOLO26m, with the same grid, time limit and markers as
  in \cref{fig:s-xgrid}.}
  \label{fig:s-xgrid-b}
\end{figure*}

\textbf{Crowded and isolated objects.} \Cref{fig:s-recall} splits the recall at IoU 0.5 into objects with a same-class
neighbour (crowded) and all others (isolated). With the pre-specified grid, fusion loses 20 to 25 points of crowded recall
against the single pass, while isolated objects gain. The refined grid recovers most of the crowded recall without losing the
gain on isolated objects. \Cref{fig:s-xgrid,fig:s-xgrid-b} show, for each configuration searched on calib, accuracy against fusion time;
the selected configurations lie on the upper edge of each cloud, and allowing more than twice the fusion time of the
pre-specified setting would have added at most 0.14\,AP on calib. Fusion times vary by about $\pm10$\,\% between
measurements, so the admissibility of variants close to the limit can change; in a repeated search, the selected
configuration and all accuracies were unchanged.

\textbf{Small objects.} \Cref{tab:s-bins} breaks the test-dev results at 2048\,px down by object size. Tiles with SCAHI improve
the bins below 16\,px for the detectors with NMS. For the NMS-free medium models YOLOv10m and YOLO26m, tiles with SCAHI do not
beat the single pass below 8\,px, and the gain of tiles appears from 8\,px upwards; this is the same head effect that the
decomposition isolates. YOLO26s gains in both bins.

\begin{table*}[tp]
\centering\small
\caption{\textbf{Small objects on test-dev} (AP for objects below 8\,px and 8--16\,px, passes at 2048\,px). Single pass with
standard post-processing; tiles with box fusion (NMS); tiles with SCAHI, pre-specified and with refined grid. Detectors of the
extension that could be trained at 2048\,px (YOLOv12m could not); small objects of YOLOv10s are in \cref{tab:s-main}.}
\label{tab:s-bins}
\begin{tabular}{@{}lcccc@{}}
\toprule
Detector & single pass & tiles, NMS & tiles, SCAHI & tiles, SCAHI refined \\
\midrule
YOLOv10m & 7.74 / 15.73 & 7.48 / 16.39 & 6.33 / 16.81 & 6.78 / 17.31 \\
YOLOv11s  & 5.14 / 12.78 & 6.33 / 13.92 & 6.79 / 15.26 & 6.88 / 15.72 \\
YOLOv11m  & 5.53 / 14.08 & 6.24 / 15.28 & 7.37 / 16.76 & 7.53 / 17.13 \\
YOLOv12s  & 4.94 / 13.69 & 6.09 / 14.65 & 6.86 / 15.91 & 7.19 / 16.39 \\
YOLO26s   & 6.17 / 14.10 & 6.29 / 14.55 & 7.10 / 15.10 & 7.54 / 15.45 \\
YOLO26m   & 7.76 / 15.93 & 8.26 / 16.18 & 7.50 / 16.78 & 7.76 / 17.21 \\
\bottomrule
\end{tabular}
\end{table*}

\begin{figure*}[tp]
  \centering
  \includegraphics[width=0.85\textwidth]{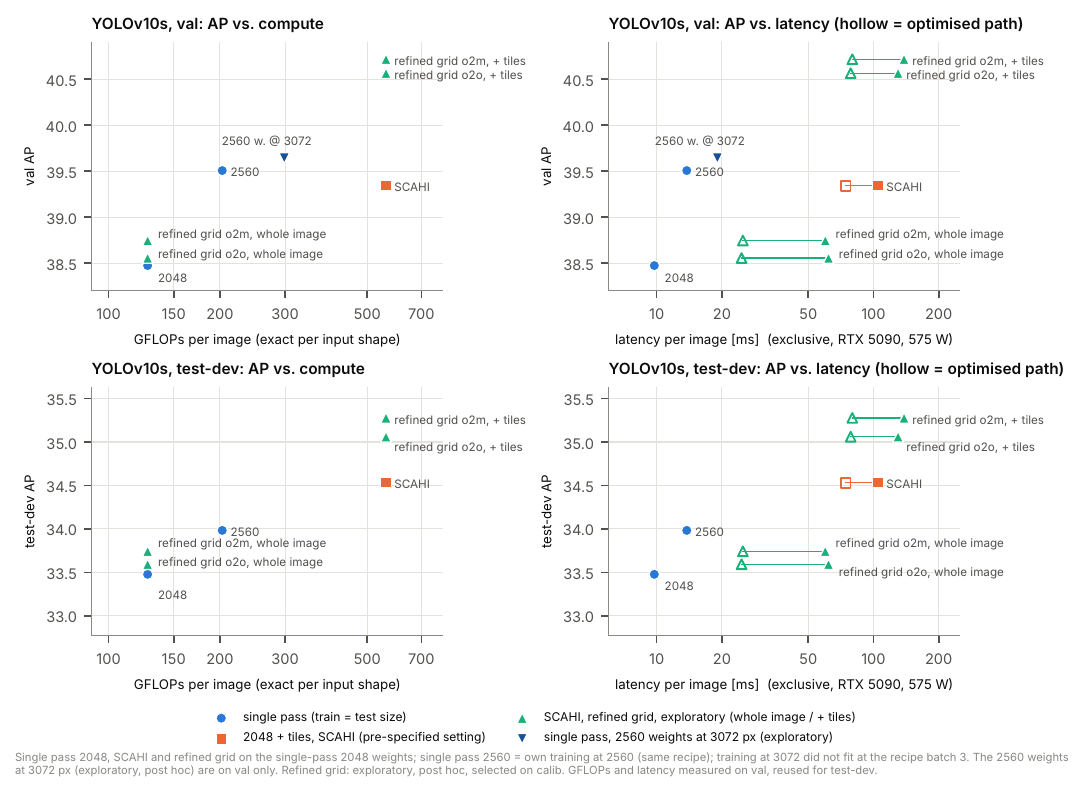}
  \caption{\textbf{Compute control for YOLOv10s}: AP against GFLOPs and against latency on val (top) and test-dev (bottom), for single passes at 2048 and 2560\,px and tiles with SCAHI (pre-specified and refined grid).}
  \label{fig:s-r3}
\end{figure*}

\textbf{Compute control.} \Cref{fig:s-r3} plots accuracy against compute and latency for YOLOv10s on both splits. The
differences to the 2560\,px single pass are, on val, $-1.03$ [$-1.36$; $-0.68$] for the 2048\,px single pass, $-0.17$
[$-0.51$; $+0.16$] for tiles with pre-specified SCAHI and $+1.21$ [$+0.90$; $+1.53$] with the refined grid; on test-dev
$-0.51$ [$-0.76$; $-0.27$], $+0.55$ [$+0.29$; $+0.79$] and $+1.30$ [$+1.05$; $+1.53$]. Compute and latency were measured on
val and apply to test-dev unchanged.

\begin{figure*}[tp]
  \centering
  \includegraphics[width=0.8\textwidth]{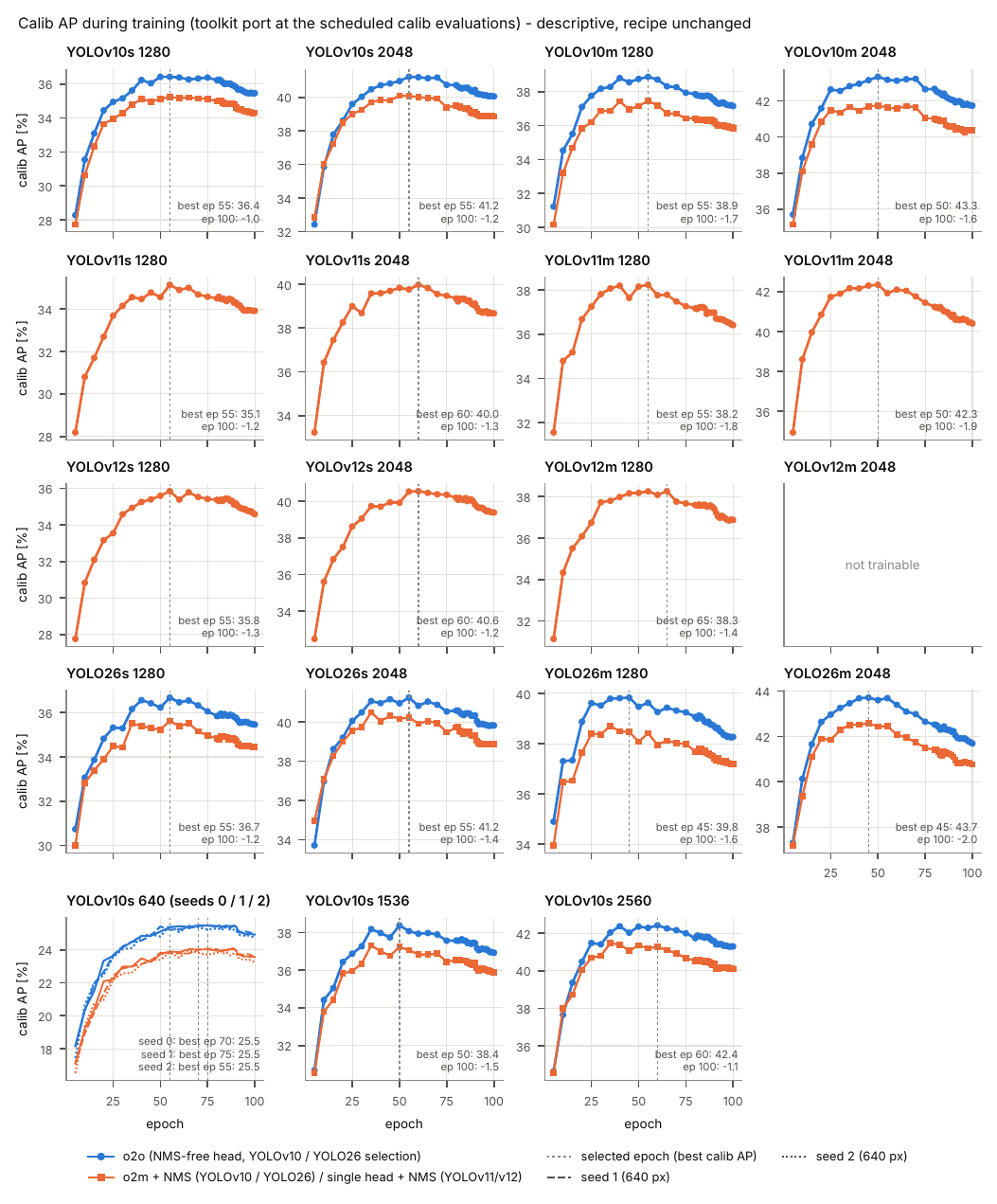}
  \caption{\textbf{Calibration AP during training} for all detectors and input sizes; the selected epoch is the maximum on calib.}
  \label{fig:s-calib}
\end{figure*}

\textbf{Training curves.} \Cref{fig:s-calib} shows the calibration AP over training for all runs, including YOLO26. The detectors peak between
epochs 45 and 75, most around epoch 50--60, and lose 0.5 to 2.0 points by epoch 100, so the selection of the epoch on calib matters; the recipe was kept unchanged
as pre-specified.

\subsection{Addendum: YOLO26s and YOLO26m}
\label{sec:s-y26}

After the verdicts of the five detectors, we added the two YOLO26 models to the extension
under a dated addendum, written before they were trained, with unchanged questions, threshold and rules; their verdicts are
reported per detector and do not enter the count over five. Both have a one-to-many and an NMS-free one-to-one branch like
YOLOv10 and regress boxes directly; they follow the same recipe (all training arguments identical apart from the model), the same
calib selection, tuning, freeze and single evaluation on val and test-dev. The selected epochs are 55 for YOLO26s and 45 for
YOLO26m at both sizes, and the search of the pre-specified SCAHI setting selected the same parameters as for the other detectors
($f=5$, five bins, thresholds 0.005 and 0.001). Their cached passes use the Ultralytics back-mapping conventions from the start;
\cref{tab:s-y26-rows} lists all 25 frozen rows per model. \Cref{tab:s-y26} gives the verdicts. Q1 and Q2 hold for both; Q3 holds for YOLO26s and is narrowly
missed by YOLO26m, as by YOLOv10m. With 20\,\% less compute (126 vs.\ 157 GFLOPs), YOLO26s at 2048\,px beats YOLO26m at 1280\,px by $+1.00$ [0.60; 1.37]
AP on val and $+0.61$ [0.34; 0.89] on test-dev. One deviation: for YOLO26s, the
single-pass rows were tuned at batch 16 and replayed at batch 1, and the replay stayed 0.08--0.09\,AP below the tuning values,
above the gate's bound of 0.05. Ultralytics pads mixed-shape batches to the full input size, which explains about 95\,\% of the
difference. Re-evaluating the confidence grid on the batch-1 detections gave the same choice and ordering, so the choice was kept
and the batch-1 values serve as reference; no rule was changed.
\begin{table}[t]
\centering\scriptsize
\setlength{\tabcolsep}{3pt}
\caption{\textbf{YOLO26 addendum} (VisDrone; $\Delta$AP with 95\,\% CI on val, $\Delta$ on test-dev; verdict on val).}
\label{tab:s-y26}
\begin{tabular}{@{}lllll@{}}
\toprule
 & \multicolumn{2}{c}{YOLO26s} & \multicolumn{2}{c}{YOLO26m} \\
\cmidrule(lr){2-3}\cmidrule(lr){4-5}
 & val & test & val & test \\
\midrule
Q1 & $+3.44$ [3.03; 3.85] & $+2.54$ & $+3.36$ [2.91; 3.83] & $+2.72$ \\
Q2, 1280\,px & $+1.57$ [1.34; 1.83] & $+2.36$ & $+1.94$ [1.65; 2.21] & $+2.24$ \\
Q2, 2048\,px & $+1.63$ [1.39; 1.85] & $+2.27$ & $+1.77$ [1.54; 1.99] & $+2.52$ \\
Q3 & $+0.90$ [0.64; 1.16] & $+1.03$ & $+0.28$ [0.04; 0.51] & $+0.51$ \\
\midrule
verdicts & \multicolumn{2}{l}{Q1, Q2, Q3 met} & \multicolumn{2}{l}{Q1, Q2 met; Q3 not met} \\
\bottomrule
\end{tabular}
\end{table}
\begin{table*}[tp]
\centering\small
\caption{\textbf{YOLO26 addendum: all frozen rows} (VisDrone, AP on val and test-dev, one evaluation each). $N$: input size of the
single pass; tiles at $z=2$. Rows marked $^\dagger$ are exploratory and enter no verdict. The questions use: Q1, single pass at
2048\,px minus 1280\,px + tiles with box fusion (one-to-one head); Q2, SCAHI minus box fusion (one-to-many head); Q3, 2048\,px +
tiles with SCAHI (one-to-many) minus the single pass at 2048\,px (one-to-one).}
\label{tab:s-y26-rows}
\begin{tabular}{@{}lllrrrr@{}}
\toprule
 & & & \multicolumn{2}{c}{YOLO26s} & \multicolumn{2}{c}{YOLO26m} \\
\cmidrule(lr){4-5}\cmidrule(lr){6-7}
$N$ & configuration & head & val & test-dev & val & test-dev \\
\midrule
1280 & single pass & one-to-one & 33.63 & 29.94 & 37.41 & 32.54 \\
 & single pass & one-to-many & 32.62 & 28.73 & 35.99 & 31.20 \\
 & SAHI & one-to-one & 31.02 & 28.51 & 34.56 & 30.51 \\
 & + tiles, box fusion & one-to-one & 34.97 & 30.61 & 37.89 & 32.58 \\
 & + tiles, box fusion & one-to-many & 34.58 & 30.09 & 37.09 & 31.98 \\
 & + tiles, SCAHI & one-to-many & 36.15 & 32.46 & 39.03 & 34.22 \\
 & + tiles, SCAHI & one-to-one & 35.90 & 31.96 & 38.86 & 33.86 \\
 & + tiles, SCAHI, refined grid$^\dagger$ & one-to-many & 37.52 & 33.12 & 40.37 & 34.92 \\
 & + tiles, SCAHI, refined grid$^\dagger$ & one-to-one & 37.08 & 32.65 & 40.19 & 34.53 \\
 & SCAHI on the single pass alone$^\dagger$ & one-to-many & 32.72 & 29.68 & 36.12 & 32.20 \\
 & SCAHI on the single pass alone$^\dagger$ & one-to-one & 32.63 & 29.42 & 36.22 & 32.05 \\
 & SCAHI on the single pass alone, refined grid$^\dagger$ & one-to-many & 34.03 & 30.34 & 37.47 & 32.87 \\
 & SCAHI on the single pass alone, refined grid$^\dagger$ & one-to-one & 33.74 & 30.06 & 37.51 & 32.68 \\
\midrule
2048 & single pass & one-to-one & 38.41 & 33.15 & 41.25 & 35.30 \\
 & single pass & one-to-many & 37.41 & 31.83 & 40.05 & 33.70 \\
 & + tiles, box fusion & one-to-one & 38.48 & 32.59 & 40.70 & 34.29 \\
 & + tiles, box fusion & one-to-many & 37.68 & 31.91 & 39.77 & 33.30 \\
 & + tiles, SCAHI & one-to-many & 39.31 & 34.18 & 41.53 & 35.82 \\
 & + tiles, SCAHI & one-to-one & 39.10 & 33.91 & 41.42 & 35.72 \\
 & + tiles, SCAHI, refined grid$^\dagger$ & one-to-many & 40.65 & 34.93 & 42.89 & 36.59 \\
 & + tiles, SCAHI, refined grid$^\dagger$ & one-to-one & 40.45 & 34.69 & 42.79 & 36.49 \\
 & SCAHI on the single pass alone$^\dagger$ & one-to-many & 37.64 & 32.83 & 40.16 & 34.73 \\
 & SCAHI on the single pass alone$^\dagger$ & one-to-one & 37.33 & 32.53 & 39.99 & 34.66 \\
 & SCAHI on the single pass alone, refined grid$^\dagger$ & one-to-many & 38.87 & 33.58 & 41.44 & 35.48 \\
 & SCAHI on the single pass alone, refined grid$^\dagger$ & one-to-one & 38.63 & 33.35 & 41.30 & 35.43 \\
\bottomrule
\end{tabular}
\end{table*}

\FloatBarrier
\section{Main study with YOLOv10s}
\label{sec:s-main}

The main study compared single passes, SAHI and multi-zoom configurations for YOLOv10s, studied five training variants
including scale-normalised training, and introduced the dense fusion. The configuration selected on calib in the main study (``selected configuration'') combines scale-normalised training,
z0.64 + z2 tiles and SCAHI with validity windows estimated on calib. \Cref{tab:s-main} lists all reference configurations;
\cref{fig:bands} relates the object sizes of val and test-dev to the sizes a pass can resolve.

\begin{figure*}[tp]
\begin{minipage}[t]{0.47\textwidth}
\centering
\begin{tikzpicture}[x=0.86cm,y=1cm,font=\scriptsize]
  \def\yb{0}
  \foreach \lo/\v/\t in {2/0.06/0.13, 3/0.25/0.23, 4/0.38/0.31, 5/0.23/0.23, 6/0.07/0.08, 7/0.01/0.01}{
     \fill[pbluel] (\lo+0.06,\yb) rectangle ++(0.42,\v*4);
     \fill[porangel!70!porange] (\lo+0.52,\yb) rectangle ++(0.42,\t*4);
  }
  \draw[->,ink2] (1,\yb) -- (9.2,\yb) node[right,align=left]{$\sigma$\\{[px]}};
  \foreach \e/\l in {1/2,2/4,3/8,4/16,5/32,6/64,7/128,8/256,9/512}{
     \draw[ink2] (\e,\yb) -- ++(0,-0.08) node[below]{\l};
  }
  \node[anchor=west] at (5.6,1.55) {\textcolor{pbluel}{$\blacksquare$} val \quad \textcolor{porangel!70!porange}{$\blacksquare$} test-dev};
  \node[anchor=west,muted] at (5.6,1.25) {GT share per size bin};
  \begin{scope}[yshift=-1.05cm]
    \fill[pblue!25] (2.585,0) -- (3.807,0.45) -- (7.585,0.45) -- (8,0) -- cycle;
    \draw[pblue,line width=0.7pt] (2.585,0) -- (3.807,0.45) -- (7.585,0.45) -- (8,0);
    \node[anchor=west] at (8.05,0.22) {full image, $z{\approx}1$};
  \end{scope}
  \begin{scope}[yshift=-1.75cm]
    \fill[paqua!25] (1.585,0) -- (2.807,0.45) -- (6.585,0.45) -- (7,0) -- cycle;
    \draw[paqua,line width=0.7pt] (1.585,0) -- (2.807,0.45) -- (6.585,0.45) -- (7,0);
    \node[anchor=west] at (7.05,0.22) {tiles, $z{=}2$};
  \end{scope}
  \draw[densely dashed,porange] (3,-1.85) -- (3,1.7) node[above,font=\tiny]{8\,px};
\end{tikzpicture}
\captionof{figure}{\textbf{Object sizes and the sizes each pass can see.} Top: share of ground truth per size bin in val and test-dev.
Bottom: the band of original object sizes $\sigma$ whose apparent size $\sigma z$ lies in the valid input range of
8--192\,px (soft edges 6--14 and 192--256\,px), for a native full-image pass and for $z{=}2$ tiles. Objects below about
4\,px are out of reach of both; test-dev has twice the share of objects below 8\,px.}
\label{fig:bands}
\end{minipage}\hfill
\begin{minipage}[t]{0.49\textwidth}
\centering\scriptsize
\setlength{\tabcolsep}{3.2pt}
\captionof{table}{\textbf{Single pass vs.\ tiling.} Toolkit AP; AP$_{<16}$: objects below 16\,px (val); MP: input megapixels per
image; latency on one RTX 5090 (val).}
\label{tab:s-main}
\begin{tabular}{@{}lrrrrr@{}}
\toprule
Configuration & AP val & AP td & AP$_{<16}$ & MP & ms \\
\midrule
single pass 640 & 22.86 & 21.82 & 6.9 & 0.25 & 4.8 \\
single pass 1280 & 33.66 & 30.25 & 14.9 & 0.94 & 6.3 \\
single pass 1536 & 35.75 & 31.60 & 17.0 & 1.33 & 7.0 \\
\textbf{single pass 2048} & \textbf{38.47} & \textbf{33.48} & 19.5 & 2.36 & 10.7 \\
SAHI, fine-tuned & 32.75 & 29.87 & 13.5 & 2.37 & 36.3 \\
SAHI (1280 weights) & 31.52 & 28.68 & 13.4 & 2.37 & 36.4 \\
1280 + tiles, box fusion & 35.00 & 30.73 & 18.5 & 9.12 & 49.9 \\
1536 + tiles, box fusion & 36.67 & 31.83 & 19.5 & 9.51 & 49.3 \\
\bottomrule
\end{tabular}

\vspace{2mm}
\centering\scriptsize
\setlength{\tabcolsep}{3pt}
\captionof{table}{\textbf{Degradation test} (recall at score $\ge0.25$, val). All 95\,\% CIs between columns are disjoint.}
\label{tab:degr}
\begin{tabular}{@{}lrrrrr@{}}
\toprule
Model & native & $d{=}2$ & small $\times2$ & $d{=}4$ & small $\times4$ \\
\midrule
single pass 1280 & 0.825 & 0.800 & 0.710 & 0.670 & 0.419 \\
single pass 2048 & 0.853 & 0.815 & 0.729 & 0.675 & --- \\
\bottomrule
\end{tabular}
\end{minipage}
\end{figure*}

\begin{figure*}[tp]
  \centering
  \begin{minipage}[t]{0.42\textwidth}
    \centering
    \includegraphics[width=\linewidth]{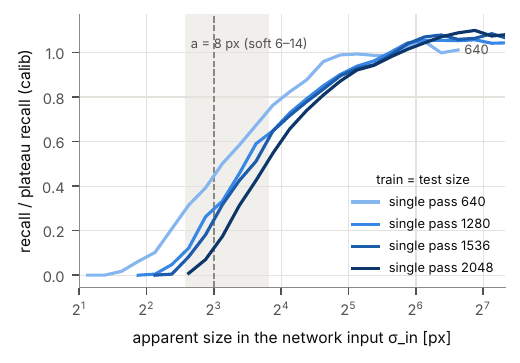}
    \caption{\textbf{Lower detection edge} on calib: recall over apparent size in the network input; 25/50/75\,\% points
    at 5.6/8.7/14.3\,px.}
    \label{fig:bandedge}
  \end{minipage}\hfill
  \begin{minipage}[t]{0.54\textwidth}
    \centering
    \includegraphics[width=\linewidth]{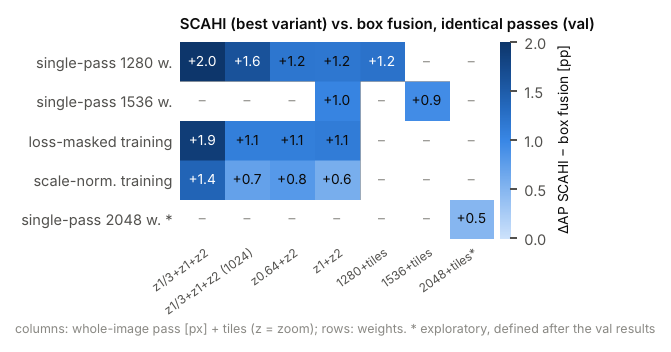}
    \caption{\textbf{Fusion gain} (best dense variant minus NMS) for every weight set and stack, configurations chosen on
    calib.}
    \label{fig:fusiongain}
  \end{minipage}
\end{figure*}

\begin{figure*}[tp]
  \centering
  \includegraphics[width=0.9\textwidth]{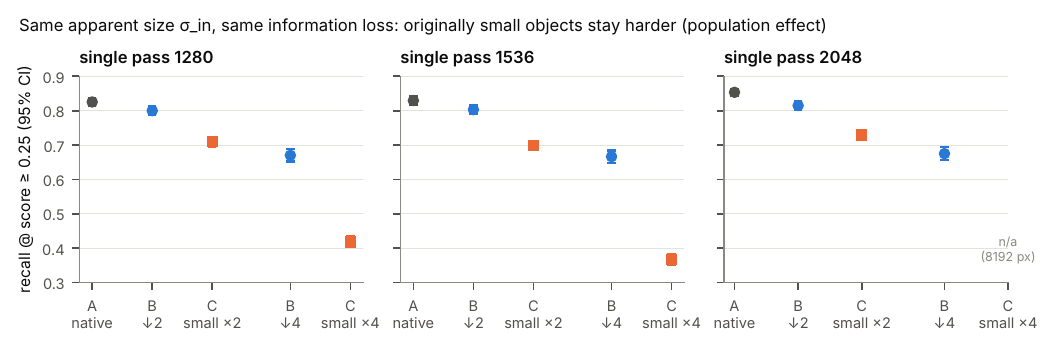}
  \caption{\textbf{Information loss vs.\ population.} Recall of large objects natively and after degradation by $d$, compared
  with originally small objects zoomed to the same apparent size (val, 95\,\% CIs).}
  \label{fig:degradation}
\end{figure*}

\textbf{Detection limit and degradation.} \Cref{fig:bandedge} shows recall over the apparent object size in the network
input on calib; it defines the lower edge of the validity windows tested for the fusion. \Cref{tab:degr} and
\cref{fig:degradation} report the degradation test: large objects that lose half of their resolution keep most of their
recall, whereas originally small objects at the same apparent size are much harder.


\begin{figure*}[tp]
  \centering
  \includegraphics[width=0.85\textwidth]{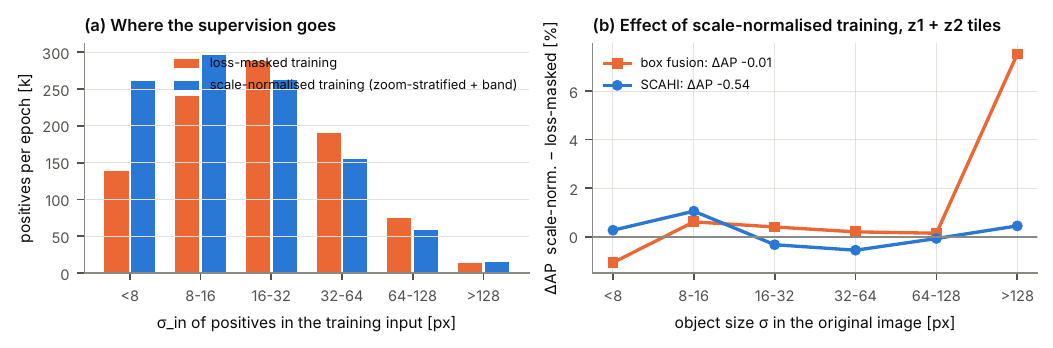}
  \caption{\textbf{H1 diagnostic}: per-size AP of T4 vs.\ T2 and positives per training chip by zoom level.}
  \label{fig:h1}
\end{figure*}

\textbf{Training variants.} \Cref{tab:train} compares the five training variants under identical inference. They differ by at
most 0.31\,AP on val. \Cref{fig:h1} explains why scale normalisation does not help: it gains on large objects and loses on small
ones, and its zoom-stratified chips carry fewer positives in the size range that matters most. \Cref{tab:s-2x2} shows that the
choice of the fusion matters more than the choice of the training variant.

\begin{table}[t]
\centering\scriptsize
\caption{\textbf{Training variants} (z1 + z2 tiles, box fusion, one-to-one head).}
\label{tab:train}
\begin{tabular}{@{}lrr@{}}
\toprule
Variant & AP val & AP test-dev \\
\midrule
ignore regions dropped & 35.76 & 31.24 \\
ignore regions painted grey & 35.76 & 31.44 \\
loss-masked training & 35.93 & 31.41 \\
loss-masked + truncation ignore & 35.62 & 31.48 \\
scale-normalised training & 35.92 & 31.69 \\
\bottomrule
\end{tabular}
\end{table}
\begin{table}[t]
\centering\scriptsize
\caption{\textbf{Training $\times$ fusion} (val AP, z0.64 + z2 tiles, one-to-many head; SCAHI with validity windows).}
\label{tab:s-2x2}
\begin{tabular}{@{}lrrr@{}}
\toprule
 & box fusion & SCAHI + windows & fusion gain \\
\midrule
loss-masked training & 34.01 & 35.33 & $+1.32$ \\
scale-normalised training & 34.82 & 35.92 & $+1.10$ \\
training gain & $+0.81$ & $+0.59$ & \\
\bottomrule
\end{tabular}
\end{table}

\begin{figure*}[tp]
  \centering
  \includegraphics[width=0.5\textwidth]{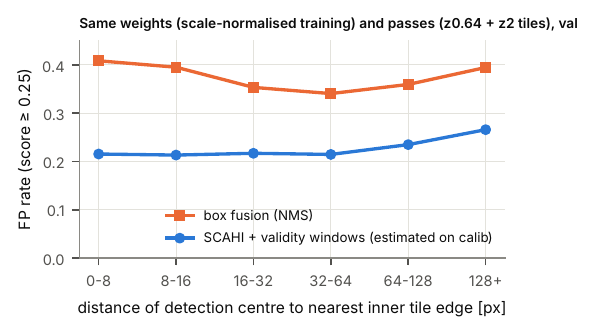}
  \caption{\textbf{False positives near tile borders} for box fusion and dense fusion on identical passes (YOLOv10s, val).}
  \label{fig:s-borderfp}
\end{figure*}

\begin{figure*}[tp]
  \centering
  \includegraphics[width=0.85\textwidth]{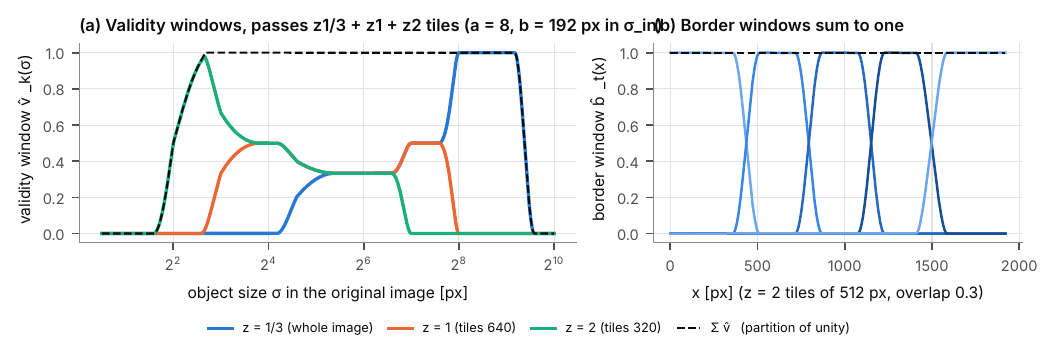}
  \caption{\textbf{Analytic windows}: border tapers (sum to 1 over the overlap) and validity windows over $\log_2(\sigma z)$
  (gate G5a: deviation from 1 at most $0.045$).}
  \label{fig:windows}
\end{figure*}

\begin{figure*}[tp]
  \centering
  \includegraphics[width=\textwidth]{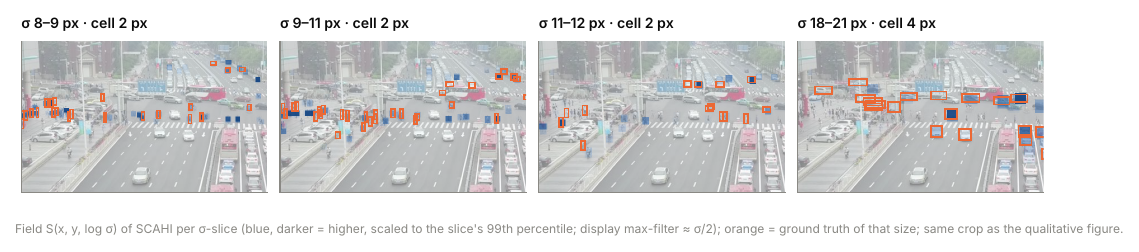}
  \caption{\textbf{Slices of the fused scale-space volume} $S(x,y,\log\sigma)$ for four size ranges (blue, darker = higher;
  orange = ground truth of that size). Small objects light up only in the small-$\sigma$ slices, larger vehicles in the
  18--21\,px slice (val image 0000295\_02000\_d\_0000031).}
  \label{fig:volume}
\end{figure*}

\begin{figure*}[tp]
  \centering
  \includegraphics[width=\textwidth]{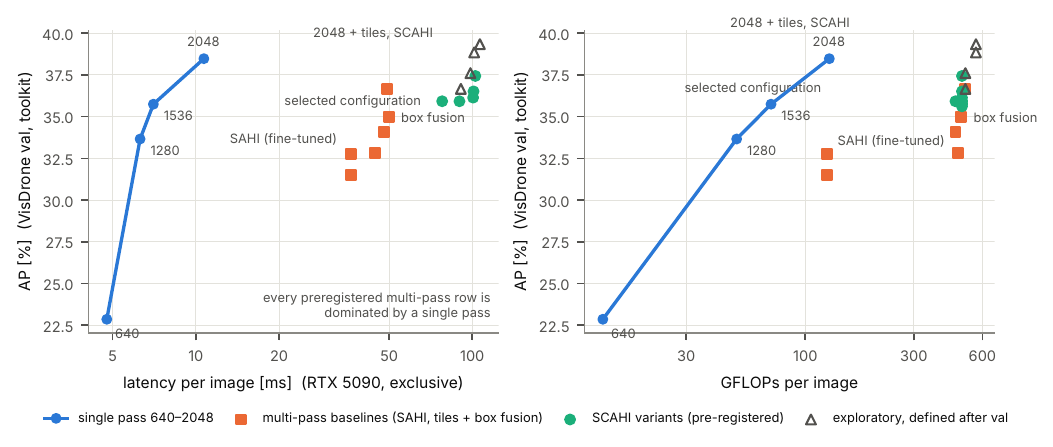}
  \caption{\textbf{Accuracy vs.\ cost} (val; final latencies with exclusive GPU and CPU). The single-pass front dominates all
  pre-specified multi-pass rows; only the exploratory 2048\,px pass with $z{=}2$ tiles and dense fusion lies above it.}
  \label{fig:s-pareto}
\end{figure*}

\textbf{Dense fusion in the main study.} \Cref{fig:s-borderfp} shows that the excess of false positives near tile borders
disappears with the dense fusion; \cref{fig:windows} shows its border and validity windows, and \cref{fig:volume} slices of the
fused volume over position and object size. \Cref{tab:s-explore} and \cref{fig:s-pareto} report the configurations that place
tiles on top of the strongest single passes, defined after the val results and confirmed on test-dev.

\begin{table}[t]
\centering\scriptsize
\setlength{\tabcolsep}{3pt}
\caption{\textbf{Exploratory rows}: single pass + $z{=}2$ tiles, $\Delta$AP vs.\ the single pass of the same weights.}
\label{tab:s-explore}
\begin{tabular}{@{}lrrr@{}}
\toprule
Rows & $\Delta$ val & $\Delta$ test-dev & AP td \\
\midrule
2048 + tiles, box fusion (o2o) & $+0.37$ & $-0.23$ & 33.25 \\
2048 + tiles, SCAHI (o2m) & $+0.87$ & $\mathbf{+1.05}$ & 34.53 \\
1536 + tiles, box fusion (o2o) & $+0.92$ & $+0.31$ & 31.92 \\
1536 + tiles, SCAHI (o2m) & $+1.86$ & $+1.92$ & 33.52 \\
\bottomrule
\end{tabular}
\end{table}

\begin{figure*}[tp]
  \centering
  \includegraphics[width=0.85\textwidth]{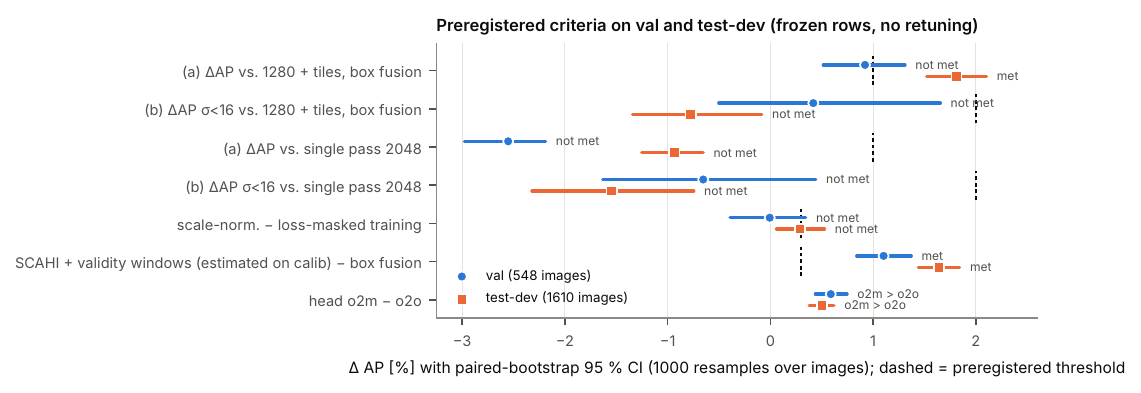}
  \caption{\textbf{Main study: pre-specified criteria on val and test-dev} with paired-bootstrap 95\,\% CIs; dashed lines mark the
  thresholds.}
  \label{fig:s-forest-main}
\end{figure*}

\begin{figure*}[tp]
  \centering
  \includegraphics[width=0.85\textwidth]{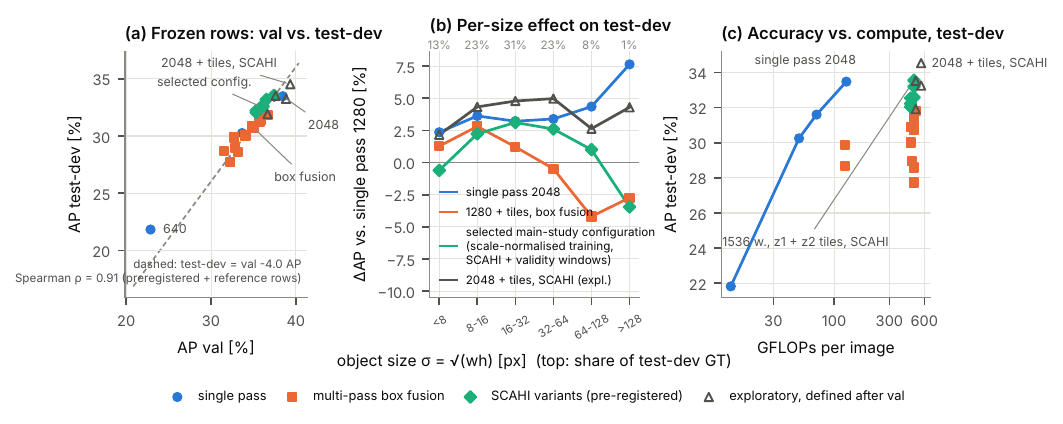}
  \caption{\textbf{val vs.\ test-dev}: AP per frozen row, per-size effect and test-dev Pareto view.}
  \label{fig:testdev}
\end{figure*}

\textbf{Pre-specified criteria and test-dev.} \Cref{tab:preregfull} and \cref{fig:s-forest-main} report the pre-specified
criteria of the main study on val and test-dev, and \cref{fig:testdev} compares all frozen configurations across the splits.

\begin{table*}[tp]
\centering\small
\caption{\textbf{Main study: pre-specified criteria with confidence intervals} (seed 0; paired bootstrap 95\,\% CI over images,
1,000 resamples; $\eps=0.30$ AP).}
\label{tab:preregfull}
\begin{tabular}{@{}lll@{}}
\toprule
Criterion & val & test-dev \\
\midrule
(a) selected config.\ $-$ (1280 + tiles, box) $\ge+1.0$ & $+0.92$ $[+0.52;+1.31]$ \textbf{not met} & $+1.81$ $[+1.52;+2.10]$ \\
(b) $\sigma<16$: $\ge+2.0$ & $+0.42$ $[-0.50;+1.65]$ not met & $-0.78$ $[-1.34;-0.08]$ \\
(c) equal AP at $\le0.7\times$ cost & not testable & not testable \\
(a) vs.\ single pass 2048 & $-2.55$ $[-2.98;-2.19]$ not met & $-0.94$ $[-1.25;-0.65]$ \\
scale-normalised $-$ loss-masked $>\eps$ & $-0.01$ $[-0.39;+0.34]$ not met & $+0.29$ $[+0.06;+0.53]$ not met \\
SCAHI $-$ box fusion $>\eps$ & $+1.10$ $[+0.84;+1.37]$ \textbf{met} & $+1.64$ $[+1.44;+1.84]$ met \\
YOLOv10 head: o2m vs.\ o2o & $+0.59$ $[+0.44;+0.75]$ & $+0.50$ $[+0.38;+0.62]$ \\
\bottomrule
\end{tabular}
\end{table*}

\FloatBarrier
\section{DOTA: protocol and detailed results}
\label{sec:s-dota}

\textbf{Data.} The DOTA study covers YOLOv10s/m, YOLOv11s, YOLOv12s and FCOS; YOLO26 was added to VisDrone only. We use DOTA v1.5
with the official horizontal-box annotations; the official boxes coincide with the minimum and
maximum of the oriented corners for every object. Train and val contain 280,196 instances. The images used are JPEG files
re-encoded at quality 70, which we report as a limitation. One training image (P1872) whose annotations lie far outside the
image was excluded from training and calibration. All objects below 8\,px and about three quarters of those between 8 and
16\,px are marked \emph{difficult}; they are used as normal positives in training, ignored by the official evaluation, and
counted in the pre-specified secondary evaluation ``all objects''.

\textbf{Evaluation.} Our port of the official DOTA devkit evaluation (horizontal boxes) reproduces the original exactly on
synthetic and realistic detections. The primary metric has no cap on detections per image; the pipeline keeps at most 200,000
detections per image, which one image reached in one configuration without changing its mAP50. A pre-specified secondary
evaluation caps detections at 4,000 per image; it, the ``all objects'' evaluation and AP@[.5:.95] lead to the same
conclusions as the primary metric.

\textbf{Training.} YOLOv10s and YOLOv10m were trained for 60 epochs at 1024 and 2048\,px, with batches of eight and three images
(effective batch 64 and 63). At 2048\,px, the CUDA allocator reported failed allocations in every epoch, which were absorbed by
retrying or by a per-image fallback of the label assignment that computes the same assignment; batch size and results are
unaffected. At 3072\,px, training did not fit even with a batch of two, because of the object density. On calib, YOLOv10m at
2048\,px stayed about 1.6 points below YOLOv10s over the whole late phase of training, whereas on DOTA-val it is ahead; the cause
remains open; possible explanations include batch normalisation with a batch of three, overfitting and the small calibration
split (141 images), which we did not test. YOLOv11s and YOLOv12s were added later under a dated addendum with unchanged hypotheses and
rules; YOLOv11s was trained at 1024 and 2048\,px, whereas YOLOv12s did not fit at 2048\,px even with a batch of two, so its
largest trainable pass is 1024\,px and the questions that require 2048\,px count as not confirmed for it.

\textbf{FCOS.} FCOS (torchvision implementation, ResNet-50-FPN, COCO weights) was added last, under a further dated addendum with
unchanged hypotheses and rules, and as a single model it is judged without correction. Its recipe was fixed in advance and is
simpler than the Ultralytics recipe of the YOLO detectors: horizontal flips only, whereas the YOLO runs used mosaic, random
scaling by up to 50\,\% and translation. FCOS was trained at 1024 and 2048\,px; at 2048\,px with a batch of one image and
gradient accumulation to the same effective batch, which the pre-specified minimum batch of two does not forbid here because
FCOS uses group normalisation in its head and frozen batch normalisation in its backbone. Every FCOS configuration uses the
detector's own score $\sqrt{p\cdot\text{centerness}}$. Two observations are specific to FCOS. First, its best inference size
on calib was 2560\,px for the 2048\,px weights (the YOLO detectors preferred their training size), plausibly because its
recipe lacks scale augmentation; the pre-specified comparison against this best single pass is reported in the main text.
Second, the search of the refined grid selected the pre-specified cell factor for FCOS at both sizes, so its refined-grid rows
equal the pre-specified ones.

\textbf{Native-resolution single pass (post hoc).} To test whether the DOTA regime merely reflects a memory limit, we ran
the frozen 2048\,px weights of YOLOv10s and YOLOv11s, which also serve the tile passes, on the whole image at native resolution
($z=1$, no tiling, no fusion), with the head, post-processing, thresholds and limits of the 2048\,px single pass. The image is
padded only at the right and bottom to the next multiple of 32, and the class offset of the batched NMS is raised from 7680 to
32\,768 so that boxes of different classes do not suppress each other at native coordinates; this changes nothing for the 2048\,px
pass. The native pass fits into the 32\,GB of the GPU for 449 and 450 of the 458 DOTA-val images (84.6\,\% of the objects); the
remaining images exceed about 30 megapixels. It takes 32\,ms per image on average and up to 29\,GB of memory. \Cref{tab:s-native}
compares it on the images that fit and, as a hybrid that falls back to tiles with SCAHI for the others, on all images. The native
pass recalls more crowded objects than tiles with SCAHI (YOLOv10s: 97.0 vs.\ 91.1\,\%) and isolated objects about as
well ($-0.7$ and $-1.3$ points), but ranks its detections worse, and it loses most on the largest objects,
which the downscaled whole-image pass of the tiled configuration covers.
\begin{table*}[tp]
\centering\small
\caption{\textbf{Native-resolution single pass on DOTA-val} (post hoc; 2048\,px weights at $z=1$). Differences in mAP50 (and
AP@[.5:.95] where stated) with paired bootstrap 95\,\% CI. Top: on the images that fit into memory (YOLOv10s 449, YOLOv11s 450
images). Middle: on all images, native where it fits and tiles + SCAHI otherwise. Bottom: native pass minus tiles + SCAHI in AP50
per object size on the images that fit.}
\label{tab:s-native}
\begin{tabular}{@{}lll@{}}
\toprule
 & YOLOv10s & YOLOv11s \\
\midrule
native $-$ single pass 2048\,px & $-2.14$ [$-4.18$; $0.02$] & $-0.95$ [$-2.75$; $1.28$] \\
native $-$ tiles + box fusion & $-3.66$ [$-6.18$; $-2.43$] & $-1.04$ [$-2.59$; $0.48$] \\
native $-$ tiles + SCAHI (refined grid) & $-5.85$ [$-8.22$; $-4.54$] & $-4.11$ [$-5.25$; $-2.43$] \\
\quad AP@[.5:.95] & $-4.92$ & $-3.47$ \\
\midrule
hybrid $-$ tiles + SCAHI (refined grid), all images & $-5.62$ [$-7.53$; $-4.49$] & $-4.02$ [$-5.12$; $-2.38$] \\
hybrid $-$ single pass 2048\,px, all images & $-0.12$ [$-2.86$; $2.25$] & $+0.59$ [$-1.53$; $3.28$] \\
\midrule
objects 8--16\,px & $-4.5$ [$-14.1$; $2.7$] & $-5.9$ [$-10.0$; $-0.8$] \\
objects 16--32\,px & $-5.3$ [$-13.8$; $-0.3$] & $-4.5$ [$-8.9$; $0.8$] \\
objects 32--64\,px & $-3.2$ [$-8.0$; $1.3$] & $+0.5$ [$-3.2$; $4.3$] \\
objects 64--128\,px & $-5.7$ [$-8.5$; $-3.4$] & $-3.9$ [$-7.3$; $-1.9$] \\
objects 128--256\,px & $-5.8$ [$-8.1$; $-1.8$] & $-4.5$ [$-8.7$; $-1.5$] \\
objects $>256$\,px & $-10.0$ [$-14.3$; $-3.4$] & $-10.8$ [$-19.4$; $-4.7$] \\
\bottomrule
\end{tabular}
\end{table*}

\textbf{Multi-zoom without tiles, and a convolutional control (post hoc).} To separate the role of tiles from that of combining
zoom levels, we fused the native whole-image pass with the 2048\,px whole-image pass by SCAHI, using the refined-grid parameters
of the detector; both passes cover the whole image, so no taper applies and $C=2$. On the images where the native pass fits
(84.6 and 84.7\,\% of the objects), this two-pass multi-zoom fusion matches tiles with SCAHI for both detectors, also for objects
above 256\,px, and it exceeds box fusion on the same two passes (\cref{tab:s-multizoom}). The information of the tiles is thus
the native-resolution detail; tiles are a way to obtain it within memory, and the fusion of zoom levels is what turns it into
accuracy. Where it fits, the two-pass fusion is also faster than tiles with SCAHI (98 and 100 against 127 and 128\,ms per image
on the same images, one exclusive run), but on the largest images it needs up to 29\,GB of GPU memory, against at most
2.5\,GB for tiles with SCAHI. As a control for the attention blocks of the YOLO models, which see a different global context at native resolution
than in training, we ran FCOS, a purely convolutional detector, as a native whole-image pass on all DOTA-val images. It shows the
same pattern: it recalls more small objects than its 2048\,px single pass but loses on the largest objects, and it stays behind
tiles with SCAHI. The native FCOS pass reaches the per-pass limit of 200{,}000 detections on 149 images and takes 446\,ms per
image; since removing the limit never raised mAP50 in \cref{tab:s-fcos-diag}(d), the limit is unlikely to disadvantage it.
\begin{table*}[tp]
\centering\small
\caption{\textbf{Multi-zoom without tiles and convolutional control} on DOTA-val (post hoc). Top: native whole-image pass plus the
2048\,px pass, fused by SCAHI (refined grid) or by box fusion, on the images where the native pass fits. Middle: FCOS as a native
whole-image pass on all images. Differences in mAP50 with paired bootstrap 95\,\% CI. Bottom: latency and memory on the images
where the native pass fits (YOLOv10s / YOLOv11s).}
\label{tab:s-multizoom}
\begin{tabular}{@{}lll@{}}
\toprule
\multicolumn{3}{@{}l}{\textit{Native + 2048\,px pass, SCAHI (multi-zoom)}} \\
 & YOLOv10s & YOLOv11s \\
\midrule
multi-zoom $-$ tiles + SCAHI & $-0.01$ [$-1.07$; $0.54$] & $+0.00$ [$-0.80$; $0.75$] \\
\quad AP@[.5:.95] & $-0.01$ & $+0.17$ \\
\quad AP50, objects $>256$\,px & $-0.25$ & $+0.62$ \\
multi-zoom $-$ single pass 2048\,px & $+3.70$ [$2.75$; $5.09$] & $+3.16$ [$1.84$; $4.32$] \\
multi-zoom $-$ native single pass & $+5.85$ [$4.36$; $7.61$] & $+4.11$ [$2.63$; $4.94$] \\
multi-zoom: SCAHI $-$ box fusion (same two passes) & $+0.85$ [$0.13$; $1.73$] & $+1.16$ [$0.14$; $1.81$] \\
multi-zoom box fusion $-$ tiles + box fusion & $+1.34$ [$-0.39$; $2.44$] & $+1.91$ [$0.80$; $2.81$] \\
\midrule
\multicolumn{3}{@{}l}{\textit{FCOS, native whole-image pass (all images)}} \\
\midrule
native $-$ single pass 2048\,px & \multicolumn{2}{l}{$+1.62$ [$-1.31$; $4.98$]; objects $>256$\,px: $-12.3$ [$-19.1$; $-0.6$] AP50} \\
native $-$ tiles + box fusion & \multicolumn{2}{l}{$-3.86$ [$-5.67$; $-1.97$]} \\
native $-$ tiles + SCAHI & \multicolumn{2}{l}{$-3.95$ [$-5.79$; $-2.10$]} \\
\midrule
\multicolumn{3}{@{}l}{\textit{Latency on the same images (ms per image, fusion on the GPU, one exclusive run)}} \\
\midrule
single pass 2048\,px & 25.4 & 24.7 \\
tiles + SCAHI & 127.1 & 127.7 \\
multi-zoom, total & 98.1 & 100.0 \\
\quad native pass / 2048\,px pass / SCAHI fusion & 64.0 / 22.6 / 11.4 & 65.0 / 23.3 / 11.7 \\
peak GPU memory, multi-zoom: max / median (GB) & 27.6 / 0.65 & 29.1 / 0.72 \\
peak GPU memory, tiles + SCAHI: max (GB) & 1.9 & 2.5 \\
\bottomrule
\end{tabular}
\end{table*}

\textbf{Why SCAHI does not exceed box fusion for FCOS.} \Cref{tab:s-fcos-diag} collects post-hoc diagnostics. Capping both
fusions at the same 200{,}000 detections per image leaves Q2 unchanged, so the uncapped output of box fusion is not the cause
(a). FCOS yields almost twice as many candidate predictions per object as the YOLO detectors (b), so a lack of redundant
evidence is not the cause either. With the coarse grid retained on calib, SCAHI recalls about 30 points fewer crowded objects
than box fusion, while isolated objects are recalled almost equally well (c): the over-merging analysed on VisDrone. Crowded
objects are only about 7\,\% of the objects on DOTA-val, which plausibly explains why selecting the cell factor by mAP50 on calib
did not favour a finer grid. Finally, the comparison is sensitive to the pipeline limit (d). The limit caps single very dense
images; without it, a few images contribute millions of low-scoring detections that rank above true positives of other images
and lower the dataset-wide precision. SCAHI then falls behind box fusion at 1024\,px, and the gain of Q3 shrinks from 5.6 to
4.1\,mAP50. The verdicts stay as computed under the pre-specified limit.
\begin{table*}[tp]
\centering\scriptsize
\caption{\textbf{FCOS diagnostics and limit check on DOTA-val} (post hoc; verdicts unchanged). (a) SCAHI (refined grid) minus box
fusion in mAP50 under the pre-specified evaluation (cap 4000), as evaluated, with both fusions capped at the top 200{,}000
detections per image, and without the pipeline limit. (b) Candidate predictions with IoU $\ge0.5$ per ground-truth object at
2048\,px, summed over all passes, mean (median). (c) Recall at IoU 0.5, SCAHI (refined grid) vs.\ box fusion. (d) mAP50 of every
affected row with and without the 200{,}000 limit (box fusion: limit per pass).}
\label{tab:s-fcos-diag}
\begin{tabular}{@{}lllll@{}}
\toprule
\multicolumn{5}{@{}l}{\textit{(a) Equal caps}} \\
Contrast & cap 4000 & as evaluated & both capped at 200k & without limit \\
\midrule
Q2, 1024\,px & $+0.38$ [$-0.94$; $1.77$] & $+0.40$ [$-0.92$; $1.79$] & $+0.40$ [$-0.92$; $1.79$] & $-1.76$ \\
Q2, 2048\,px & $+0.10$ [$-0.96$; $1.33$] & $+0.09$ [$-0.91$; $1.37$] & $+0.09$ [$-0.91$; $1.36$] & $+0.14$ \\
\midrule
\multicolumn{5}{@{}l}{\textit{(b) Redundancy: candidates per object, all / crowded / isolated}} \\
FCOS & \multicolumn{4}{l}{62.7 (44) / 95.4 (85) / 60.1 (41)} \\
YOLOv10s & \multicolumn{4}{l}{35.5 (33) / 47.6 (42) / 34.5 (32)} \\
YOLOv11s & \multicolumn{4}{l}{35.8 (33) / 48.1 (43) / 34.8 (32)} \\
\midrule
\multicolumn{5}{@{}l}{\textit{(c) Recall, SCAHI vs.\ box fusion}} \\
 & \multicolumn{2}{l}{crowded (2515 objects)} & \multicolumn{2}{l}{isolated (31\,489 objects)} \\
1024\,px & \multicolumn{2}{l}{61.1 vs.\ 91.5: $-30.3$ [$-35.2$; $-24.4$]} & \multicolumn{2}{l}{92.5 vs.\ 93.4: $-0.9$ [$-1.6$; $-0.2$]} \\
2048\,px & \multicolumn{2}{l}{62.6 vs.\ 93.4: $-30.9$ [$-33.5$; $-27.1$]} & \multicolumn{2}{l}{93.9 vs.\ 95.3: $-1.4$ [$-2.2$; $-0.7$]} \\
\midrule
\multicolumn{5}{@{}l}{\textit{(d) Limit check}} \\
Row & affected images & with limit & without limit & $\Delta$ \\
\midrule
single pass, 1024\,px & 24 & 42.64 & 42.64 & $\phantom{-}0.00$ \\
tiles + box fusion, 1024\,px & 30 & 52.56 & 52.56 & $\phantom{-}0.00$ \\
tiles + SCAHI, 1024\,px & 8 & 52.96 & 50.81 & $-2.16$ \\
single pass, 2048\,px & 76 & 55.01 & 55.01 & $\phantom{-}0.00$ \\
single pass, 2048\,px weights at 2560\,px & 82 & 58.12 & 58.12 & $\phantom{-}0.00$ \\
tiles + box fusion, 2048\,px & 78 & 60.49 & 58.96 & $-1.53$ \\
tiles + SCAHI, 2048\,px & 1 & 60.59 & 59.10 & $-1.48$ \\
\bottomrule
\end{tabular}
\end{table*}

\textbf{All configurations and secondary evaluations.} \Cref{tab:s-dota-all} lists all frozen configurations of the five
detectors under the primary and the three secondary evaluations, and \cref{tab:s-dota-dh3} the decisive comparison of tiles with
SCAHI against the single pass under all four. The gain holds under every evaluation and grows when difficult objects are
counted. \Cref{tab:s-dota-class} breaks the result down by class.
\begin{table*}[tp]
\centering\scriptsize
\caption{\textbf{All frozen configurations on DOTA-val} for the five detectors (devkit mAP50 without cap, with cap 4000, with
difficult objects counted, and AP@[.5:.95]). YOLOv10: single pass and SAHI on the one-to-one head, box fusion and SCAHI on the
one-to-many head unless marked; YOLOv11s/v12s and FCOS use their single head (FCOS score $\sqrt{p\cdot\text{centerness}}$). For box
fusion, the pipeline limit applies per pass and the merged output is not capped.}
\label{tab:s-dota-all}
\begin{tabular}{@{}lllrrrr@{}}
\toprule
Detector & $N$ & configuration & mAP50 & cap 4000 & all objects & AP@[.5:.95] \\
\midrule
YOLOv10s & 1024 & single pass & 55.80 & 55.76 & 50.59 & 36.14 \\
 &  & tiles + box fusion (o2o head) & 63.18 & 63.15 & 60.39 & 41.23 \\
 &  & tiles + box fusion & 64.86 & 64.63 & 61.53 & 43.50 \\
 &  & tiles + SCAHI & 66.17 & 66.02 & 62.65 & 44.75 \\
 &  & tiles + SCAHI, refined grid & 66.17 & 65.90 & 62.71 & 45.63 \\
 &  & SCAHI, single pass (no tiles) & 56.29 & 56.30 & 50.94 & 36.44 \\
 &  & SCAHI, single pass (no tiles), refined grid & 57.22 & 57.18 & 51.88 & 37.31 \\
 &  & SAHI & 59.20 & 59.11 & 56.31 & 37.46 \\
 & 2048 & single pass & 65.50 & 65.27 & 60.71 & 44.65 \\
 &  & single pass, 2048 weights at 2560\,px (crossing) & 65.50 & 65.27 & 60.71 & 44.65 \\
 &  & tiles + box fusion (o2o head) & 68.29 & 68.33 & 64.43 & 46.85 \\
 &  & tiles + box fusion & 68.46 & 68.49 & 64.56 & 47.41 \\
 &  & tiles + SCAHI & 70.64 & 70.65 & 66.69 & 49.15 \\
 &  & tiles + SCAHI, refined grid & 71.00 & 71.06 & 67.03 & 50.13 \\
 &  & SCAHI, single pass (no tiles) & 66.58 & 66.13 & 61.18 & 44.87 \\
 &  & SCAHI, single pass (no tiles), refined grid & 67.27 & 66.96 & 62.04 & 45.94 \\
\midrule
YOLOv10m & 1024 & single pass & 58.44 & 58.42 & 52.82 & 38.37 \\
 &  & tiles + box fusion (o2o head) & 63.97 & 63.82 & 61.17 & 43.09 \\
 &  & tiles + box fusion & 64.57 & 64.48 & 61.37 & 43.87 \\
 &  & tiles + SCAHI & 66.69 & 66.54 & 63.21 & 46.33 \\
 &  & tiles + SCAHI, refined grid & 67.34 & 67.08 & 63.88 & 47.02 \\
 &  & SCAHI, single pass (no tiles) & 58.61 & 58.61 & 52.98 & 38.74 \\
 &  & SCAHI, single pass (no tiles), refined grid & 59.29 & 59.26 & 53.60 & 39.50 \\
 &  & SAHI & 60.96 & 60.87 & 58.04 & 39.78 \\
 & 2048 & single pass & 67.39 & 66.99 & 63.06 & 46.13 \\
 &  & single pass, 2048 weights at 2560\,px (crossing) & 67.39 & 66.99 & 63.06 & 46.13 \\
 &  & tiles + box fusion (o2o head) & 68.71 & 68.54 & 65.63 & 47.74 \\
 &  & tiles + box fusion & 68.13 & 68.08 & 64.61 & 47.85 \\
 &  & tiles + SCAHI & 71.32 & 71.08 & 67.64 & 50.57 \\
 &  & tiles + SCAHI, refined grid & 71.73 & 71.44 & 68.10 & 51.30 \\
 &  & SCAHI, single pass (no tiles) & 67.67 & 67.20 & 62.80 & 46.48 \\
 &  & SCAHI, single pass (no tiles), refined grid & 68.42 & 68.02 & 63.52 & 47.30 \\
\midrule
YOLOv11s & 1024 & single pass & 56.16 & 56.16 & 50.91 & 36.30 \\
 &  & tiles + box fusion & 62.14 & 61.92 & 59.24 & 41.06 \\
 &  & tiles + SCAHI & 65.20 & 65.07 & 61.56 & 43.85 \\
 &  & tiles + SCAHI, refined grid & 65.26 & 65.09 & 61.67 & 44.61 \\
 &  & SCAHI, single pass (no tiles) & 55.68 & 55.64 & 50.54 & 35.76 \\
 &  & SCAHI, single pass (no tiles), refined grid & 56.47 & 56.46 & 51.34 & 36.45 \\
 & 2048 & single pass & 65.19 & 64.98 & 60.28 & 44.69 \\
 &  & single pass, 2048 weights at 2560\,px (crossing) & 65.19 & 64.98 & 60.28 & 44.69 \\
 &  & tiles + box fusion & 66.68 & 66.49 & 62.91 & 45.93 \\
 &  & tiles + SCAHI & 69.14 & 68.97 & 65.38 & 48.42 \\
 &  & tiles + SCAHI, refined grid & 69.81 & 69.66 & 66.01 & 49.36 \\
 &  & SCAHI, single pass (no tiles) & 64.92 & 64.71 & 59.74 & 44.02 \\
 &  & SCAHI, single pass (no tiles), refined grid & 65.45 & 65.24 & 60.62 & 44.94 \\
\midrule
YOLOv12s & 1024 & single pass & 56.36 & 56.36 & 51.27 & 36.62 \\
 &  & tiles + box fusion & 62.88 & 62.79 & 59.75 & 42.17 \\
 &  & tiles + SCAHI & 65.18 & 65.07 & 61.80 & 44.16 \\
 &  & tiles + SCAHI, refined grid & 65.97 & 65.88 & 62.65 & 45.31 \\
 &  & SCAHI, single pass (no tiles) & 55.78 & 55.69 & 50.50 & 35.92 \\
 &  & SCAHI, single pass (no tiles), refined grid & 56.74 & 56.69 & 51.70 & 36.95 \\
\midrule
FCOS & 1024 & single pass & 42.64 & 42.63 & 38.60 & 23.97 \\
 &  & tiles + box fusion & 52.56 & 52.56 & 48.70 & 31.45 \\
 &  & tiles + SCAHI & 52.96 & 52.94 & 48.94 & 31.99 \\
 &  & tiles + SCAHI, refined grid & 52.96 & 52.94 & 48.94 & 31.99 \\
 &  & SCAHI, single pass (no tiles) & 41.89 & 41.89 & 37.81 & 22.97 \\
 &  & SCAHI, single pass (no tiles), refined grid & 42.45 & 42.44 & 38.49 & 23.74 \\
 & 2048 & single pass & 55.01 & 54.97 & 49.58 & 33.91 \\
 &  & single pass, 2048 weights at 2560\,px (crossing) & 58.12 & 58.12 & 52.67 & 36.00 \\
 &  & tiles + box fusion & 60.49 & 60.45 & 56.21 & 37.50 \\
 &  & tiles + SCAHI & 60.59 & 60.55 & 55.79 & 37.55 \\
 &  & tiles + SCAHI, refined grid & 60.59 & 60.55 & 55.79 & 37.55 \\
 &  & SCAHI, single pass (no tiles) & 54.04 & 54.04 & 48.50 & 32.21 \\
 &  & SCAHI, single pass (no tiles), refined grid & 54.04 & 54.04 & 48.50 & 32.21 \\
\bottomrule
\end{tabular}
\end{table*}

\begin{table*}[tp]
\centering\small
\caption{\textbf{Q3 on DOTA-val under four evaluations}: tiles + SCAHI (refined grid) vs.\ the single pass, both at the largest
trainable size, with paired bootstrap 95\,\% CI. Only the primary metric carries the pre-specified verdict. $^\dagger$YOLOv12s
could only be trained at 1024\,px, where the single pass is weaker; not comparable with the 2048\,px values.}
\label{tab:s-dota-dh3}
\begin{tabular}{@{}lllrrl@{}}
\toprule
Detector & $N$ & evaluation & single pass & tiles + SCAHI & $\Delta$ [95\,\% CI] \\
\midrule
YOLOv10s & 2048 & mAP50 (primary) & 65.50 & 71.00 & $+5.50$ [+3.55; +7.77] \\
 &  & mAP50, all objects & 60.71 & 67.03 & $+6.32$ [+4.54; +8.64] \\
 &  & mAP50, cap 4000 & 65.27 & 71.06 & $+5.79$ [+3.64; +8.18] \\
 &  & AP@[.5:.95] & 44.65 & 50.13 & $+5.48$ [+4.17; +7.03] \\
\midrule
YOLOv10m & 2048 & mAP50 (primary) & 67.39 & 71.73 & $+4.34$ [+3.02; +6.19] \\
 &  & mAP50, all objects & 63.06 & 68.10 & $+5.04$ [+3.63; +6.87] \\
 &  & mAP50, cap 4000 & 66.99 & 71.44 & $+4.45$ [+3.07; +6.24] \\
 &  & AP@[.5:.95] & 46.13 & 51.30 & $+5.17$ [+3.84; +6.50] \\
\midrule
YOLOv11s & 2048 & mAP50 (primary) & 65.19 & 69.81 & $+4.62$ [+2.76; +6.62] \\
 &  & mAP50, all objects & 60.28 & 66.01 & $+5.73$ [+3.97; +7.92] \\
 &  & mAP50, cap 4000 & 64.98 & 69.66 & $+4.68$ [+2.76; +6.66] \\
 &  & AP@[.5:.95] & 44.69 & 49.36 & $+4.67$ [+3.43; +6.20] \\
\midrule
YOLOv12s & 1024$^\dagger$ & mAP50 (primary) & 56.36 & 65.97 & $+9.62$ [+7.10; +12.22] \\
 &  & mAP50, all objects & 51.27 & 62.65 & $+11.39$ [+9.09; +13.96] \\
 &  & mAP50, cap 4000 & 56.36 & 65.88 & $+9.52$ [+7.03; +11.89] \\
 &  & AP@[.5:.95] & 36.62 & 45.31 & $+8.69$ [+6.74; +10.60] \\
\midrule
FCOS & 2048 & mAP50 (primary) & 55.01 & 60.59 & $+5.57$ [+3.77; +7.65] \\
 &  & mAP50, all objects & 49.58 & 55.79 & $+6.21$ [+4.25; +8.03] \\
 &  & mAP50, cap 4000 & 54.97 & 60.55 & $+5.58$ [+3.71; +7.62] \\
 &  & AP@[.5:.95] & 33.91 & 37.55 & $+3.64$ [+2.38; +4.94] \\
\bottomrule
\end{tabular}
\end{table*}

\begin{table*}[tp]
\centering\scriptsize
\caption{\textbf{Per-class AP50 on DOTA-val} at the largest trainable input size $N$: single pass (SP), tiles + box fusion (BF), tiles + SCAHI
with refined grid (SC).}
\label{tab:s-dota-class}
\resizebox{\textwidth}{!}{%
\begin{tabular}{@{}lrrrrrrrrrrrrrrr@{}}
\toprule
 & \multicolumn{3}{c}{YOLOv10s (2048\,px)} & \multicolumn{3}{c}{YOLOv10m (2048\,px)} & \multicolumn{3}{c}{YOLOv11s (2048\,px)} & \multicolumn{3}{c}{YOLOv12s (1024\,px)} & \multicolumn{3}{c}{FCOS (2048\,px)} \\
\cmidrule(lr){2-4}\cmidrule(lr){5-7}\cmidrule(lr){8-10}\cmidrule(lr){11-13}\cmidrule(lr){14-16}
class & SP & BF & SC & SP & BF & SC & SP & BF & SC & SP & BF & SC & SP & BF & SC \\
\midrule
plane & 88.6 & 90.2 & 90.3 & 89.9 & 90.3 & 90.4 & 89.3 & 90.0 & 90.1 & 80.6 & 89.7 & 89.9 & 79.7 & 88.9 & 89.0 \\
baseball-diamond & 70.4 & 64.8 & 71.5 & 72.3 & 70.1 & 75.2 & 71.5 & 70.9 & 72.5 & 66.1 & 60.9 & 66.8 & 60.1 & 60.5 & 64.9 \\
bridge & 45.5 & 48.0 & 51.1 & 42.7 & 43.9 & 47.2 & 45.2 & 44.7 & 48.6 & 27.7 & 35.6 & 39.6 & 32.6 & 38.2 & 41.4 \\
ground-track-field & 59.2 & 68.4 & 70.6 & 69.0 & 67.9 & 73.3 & 59.2 & 62.4 & 67.7 & 47.3 & 58.6 & 60.6 & 43.8 & 45.6 & 41.7 \\
small-vehicle & 70.7 & 77.7 & 81.7 & 75.2 & 80.5 & 82.9 & 72.7 & 79.1 & 82.2 & 58.1 & 75.3 & 78.9 & 57.8 & 67.9 & 68.7 \\
large-vehicle & 79.9 & 81.7 & 82.5 & 80.6 & 76.7 & 83.8 & 79.9 & 81.7 & 84.1 & 77.0 & 78.0 & 81.5 & 76.3 & 78.0 & 71.6 \\
ship & 88.5 & 89.7 & 89.9 & 89.4 & 89.4 & 89.8 & 88.9 & 89.2 & 89.7 & 78.6 & 89.2 & 89.2 & 78.3 & 86.0 & 80.7 \\
tennis-court & 90.7 & 90.7 & 90.9 & 90.7 & 90.8 & 90.9 & 90.8 & 90.8 & 90.9 & 90.6 & 90.7 & 90.8 & 90.0 & 90.5 & 90.7 \\
basketball-court & 52.3 & 61.7 & 63.2 & 58.3 & 67.7 & 69.8 & 54.2 & 59.5 & 60.8 & 47.2 & 48.5 & 54.9 & 43.1 & 41.6 & 45.6 \\
storage-tank & 81.4 & 86.1 & 86.6 & 82.0 & 85.9 & 87.1 & 83.5 & 86.8 & 87.4 & 62.7 & 76.8 & 79.6 & 68.9 & 81.4 & 81.0 \\
soccer-ball-field & 65.8 & 76.4 & 74.2 & 63.3 & 66.2 & 72.4 & 61.4 & 63.3 & 70.6 & 63.3 & 67.3 & 71.6 & 56.9 & 66.3 & 66.6 \\
roundabout & 54.8 & 66.0 & 67.9 & 58.9 & 67.6 & 69.2 & 58.1 & 65.2 & 68.7 & 33.8 & 56.2 & 59.1 & 49.0 & 61.5 & 61.7 \\
harbor & 82.4 & 80.8 & 85.7 & 83.7 & 77.6 & 86.2 & 83.9 & 77.7 & 85.9 & 78.3 & 75.2 & 82.8 & 69.7 & 70.7 & 73.6 \\
swimming-pool & 69.5 & 75.6 & 78.3 & 71.1 & 70.1 & 76.6 & 68.4 & 75.4 & 79.5 & 60.8 & 72.0 & 72.6 & 55.5 & 63.9 & 65.2 \\
helicopter & 48.1 & 35.3 & 49.5 & 50.8 & 44.8 & 52.4 & 35.5 & 29.7 & 37.7 & 29.3 & 32.0 & 37.6 & 18.4 & 26.9 & 27.0 \\
container-crane & 0.1 & 2.3 & 2.2 & 0.4 & 0.8 & 0.5 & 0.5 & 0.4 & 0.6 & 0.1 & 0.0 & 0.2 & 0.0 & 0.0 & 0.0 \\
\midrule
mAP50 & 65.5 & 68.5 & 71.0 & 67.4 & 68.1 & 71.7 & 65.2 & 66.7 & 69.8 & 56.4 & 62.9 & 66.0 & 55.0 & 60.5 & 60.6 \\
\bottomrule
\end{tabular}}
\end{table*}

\textbf{Latency.} On DOTA (\cref{tab:s-latency-dota}), the tile passes account for 90--93\,\% of the total time when the
fusion runs on the GPU (optimised path); SCAHI adds about 13\,ms per image, and tiles with SCAHI are faster than SAHI for both detectors where
SAHI was run. The GPU fusion processes the image in spatial blocks with a margin derived from the largest cell of the peak grid;
on every image of DOTA-val and VisDrone val, the reference GPU implementation reproduces the CPU reference exactly (no differing
detection among 42 million), and the optimised path used for the latencies matches it up to float32 rounding.
\Cref{tab:s-latency} gives the corresponding breakdown for VisDrone.

\textbf{SCAHI as a replacement for SAHI's merge (Q4).} For YOLOv10s/m, this comparison uses exactly the passes of the frozen SAHI
configuration (1024\,px weights, SAHI's 640\,px slices and its standard full-image pass at 640\,px). The caches reproduce SAHI's own inference
path, including the fused model, one slice per forward pass and Ultralytics' resize and back-mapping; recomputing SAHI from
them gives its frozen calib mAP50 exactly. SCAHI was tuned on calib as for all other rows. For YOLOv11s and YOLOv12s, which were added under a later addendum, SAHI was first
tuned on calib with the same procedure (both select an overlap of 0.2 and NMS), and SCAHI was then tuned on exactly these passes;
the two pairs were decided separately. \Cref{tab:s-dh5} reports the results; the gain holds under every evaluation and for all
four detectors. Because YOLOv11s and YOLOv12s have a single dense head, SAHI and SCAHI see the same predictions there, which
isolates the merge step. By object size, SCAHI improves on SAHI in all bins above 8\,px, most for
objects above 128\,px (4.4 and 4.6\,AP50), which SAHI's slices cut into fragments. With the pre-specified setting, SCAHI lowers
the recall of crowded objects for every detector (e.g.\ 67 vs.\ 91\,\% for YOLOv10s, and by 17 and 19 points for YOLOv11s and
YOLOv12s); the refined grid restores it and lies above SAHI, as on the other configurations. With the fusion on the GPU, SAHI's passes
followed by SCAHI take 147 and 219\,ms per image for YOLOv10s and YOLOv10m, and SAHI itself 195 and 268\,ms in the same run.
The difference in the passes (about 45\,ms) is overhead of the SAHI library on the CPU, mainly image loading and per-detection
Python objects; it is not a property of SCAHI. SCAHI's merge takes about 12\,ms on the GPU, SAHI's merge 14--16\,ms on the CPU.
\begin{table*}[tp]
\centering\small
\setlength{\tabcolsep}{4pt}
\caption{\textbf{SCAHI on SAHI's passes (Q4)} on DOTA-val (mAP50 without cap unless stated). Differences to SAHI with paired
bootstrap 95\,\% CI; Q4 is decided on the primary metric for the refined grid, Holm over YOLOv10s/m and, separately, over
YOLOv11s/12s. Latency (ms per image, fusion on the GPU, optimised path) from one exclusive run; not measured for YOLOv11s/12s.}
\label{tab:s-dh5}
\begin{tabular}{@{}lllll@{}}
\toprule
 & YOLOv10s & YOLOv10m & YOLOv11s & YOLOv12s \\
\midrule
SAHI & 59.20 & 60.96 & 57.90 & 60.21 \\
SAHI passes + SCAHI & 61.96 & 62.98 & 61.36 & 63.13 \\
SAHI passes + SCAHI, refined grid & 62.08 & 63.22 & 61.87 & 63.30 \\
\midrule
$\Delta$ refined grid (primary, Q4) & $+2.87$ [$2.13$; $4.33$] & $+2.26$ [$1.08$; $3.25$] & $+3.97$ [$2.93$; $5.06$] & $+3.10$ [$2.23$; $4.08$] \\
$\Delta$ refined grid, cap 4000 & $+2.80$ [$2.07$; $4.27$] & $+2.13$ [$0.86$; $3.16$] & $+4.05$ [$2.91$; $5.06$] & $+3.02$ [$2.17$; $4.03$] \\
$\Delta$ refined grid, all objects & $+2.67$ [$1.97$; $3.98$] & $+1.99$ [$1.11$; $2.83$] & $+3.92$ [$2.84$; $4.79$] & $+2.99$ [$2.21$; $3.95$] \\
$\Delta$ refined grid, AP@[.5:.95] & $+4.15$ [$3.24$; $5.18$] & $+3.73$ [$2.91$; $4.57$] & $+4.25$ [$3.50$; $4.99$] & $+3.40$ [$2.74$; $4.13$] \\
$\Delta$ pre-specified setting (primary) & $+2.76$ [$1.65$; $4.12$] & $+2.02$ [$0.83$; $3.06$] & $+3.46$ [$2.38$; $4.56$] & $+2.92$ [$1.92$; $3.90$] \\
\midrule
latency SAHI / SAHI passes + SCAHI & 195 / 147 & 268 / 219 & -- & -- \\
\bottomrule
\end{tabular}
\end{table*}
\begin{table*}[tp]
\centering\small
\caption{\textbf{Latency on DOTA-val} (ms per image, mean over 458 images; RTX 5090 at default power, exclusive use, batch 1;
all rows from one run on the same machine as Tab.~\ref{tab:compute}). Totals including the tile passes. Box fusion and SCAHI on the
CPU use the reference passes (tile passes 137.9, 188.7, 125.9 and 133.8\,ms); SCAHI on the GPU uses the optimised path of
Tab.~\ref{tab:compute} (passes 120.4, 172.2, 121.0 and 129.6\,ms; fusion 13.2--13.8\,ms). A one-to-one matching of its detections
with the CPU reference shows no deviation beyond float32 rounding. Box fusion runs on the CPU, so the table shows how the cost
splits, not which merge step is faster. Peak GPU memory of the SCAHI fusion: 1.6--1.9\,GB.}
\label{tab:s-latency-dota}
\begin{tabular}{@{}lrrrrr@{}}
\toprule
Detector & single pass & tiles + box fusion & tiles + SCAHI (CPU) & tiles + SCAHI (GPU) & SAHI \\
\midrule
YOLOv10s, 2048\,px & 28.0 & 175.5 & 300.9 & 133.6 & 194.6 \\
YOLOv10m, 2048\,px & 34.7 & 225.2 & 398.0 & 186.0 & 268.1 \\
YOLOv11s, 2048\,px & 28.5 & 165.3 & 286.9 & 134.2 & -- \\
YOLOv12s, 1024\,px & 21.7 & 167.1 & 334.0 & 142.9 & -- \\
\bottomrule
\end{tabular}
\end{table*}

\begin{table*}[tp]
\centering\small
\caption{\textbf{Latency breakdown on VisDrone} (ms per image, YOLOv10s, exclusive GPU and CPU, RTX 5090 at default power,
batch 1, one run): time of the passes and of the fusion on the CPU or the GPU.}
\label{tab:s-latency}
\begin{tabular}{@{}lllrrrrr@{}}
\toprule
Dataset & Detector & configuration & passes & fusion CPU & fusion GPU & total (CPU) & total (GPU) \\
\midrule
VisDrone val & YOLOv10s & single pass 2048 & 10.3 & -- & -- & 10.3 & 10.3 \\
 & YOLOv10s & single pass 2560 & 14.4 & -- & -- & 14.4 & 14.4 \\
 & YOLOv10s & tiles + SCAHI (pre-specified) & 75.6 & 30.8 & 9.4 & 106.4 & 85.0 \\
 & YOLOv10s & tiles + SCAHI, refined grid & 76.6 & 60.1 & 10.2 & 136.7 & 86.8 \\
\bottomrule
\end{tabular}
\end{table*}

\begin{figure*}[tp]
  \centering
  \includegraphics[width=0.85\textwidth]{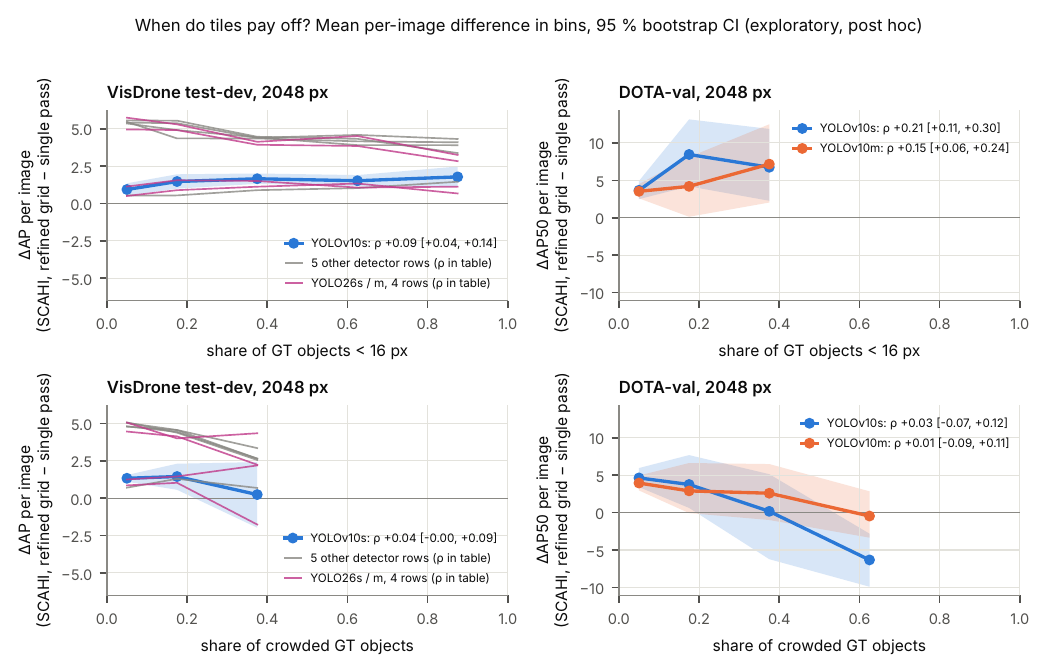}
  \caption{\textbf{When do tiles pay off?} Per-image gain of tiles with SCAHI (refined grid) over the single pass at 2048\,px, in bins of the share of objects below 16\,px and of crowded objects, on VisDrone test-dev and DOTA-val (post hoc; mean with 95\,\% CI).}
  \label{fig:s-payoff}
\end{figure*}

\textbf{When do tiles pay off?} \Cref{fig:s-payoff} relates the per-image gain of tiles with SCAHI (refined grid) over the
single pass at 2048\,px to the share of objects below 16\,px and to the share of crowded objects (post hoc). On DOTA, the gain
increases with the share of small objects (Spearman $\rho=0.21$ and $0.15$ for YOLOv10s/m, both intervals above zero); on
VisDrone test-dev the relation is weak and changes sign between detectors and heads ($-0.10$ to $+0.09$, including YOLO26). The
relation with the share of crowded objects is weak at most in both datasets (up to $+0.08$ for YOLO26s on VisDrone).

\begin{figure*}[tp]
  \centering
  \includegraphics[width=0.6\textwidth]{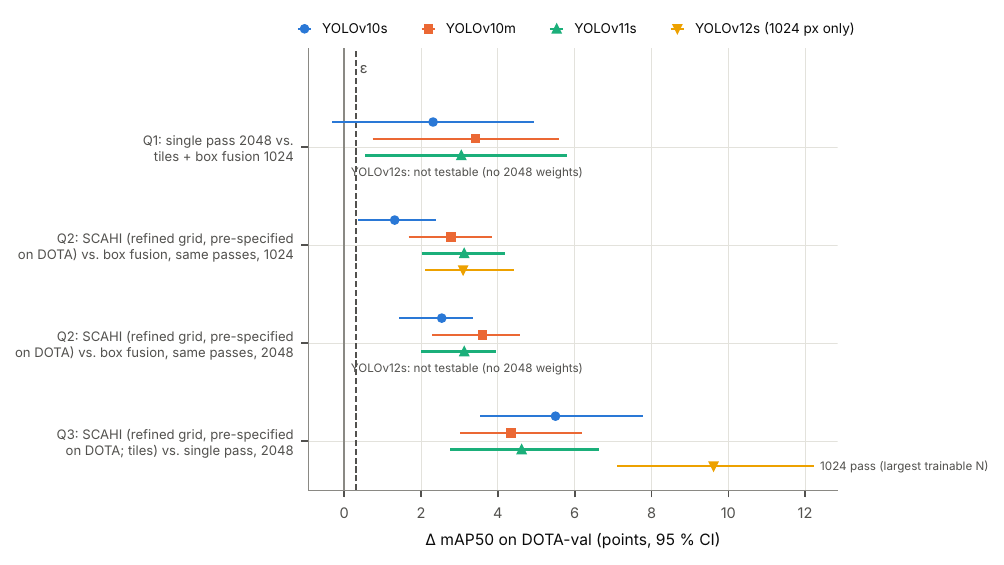}
  \caption{\textbf{Pre-specified questions on DOTA-val}: differences in mAP50 with 95\,\% confidence intervals; dashed: $\eps$.}
  \label{fig:s-dota-q}
\end{figure*}

\begin{figure*}[tp]
  \centering
  \includegraphics[width=0.8\textwidth]{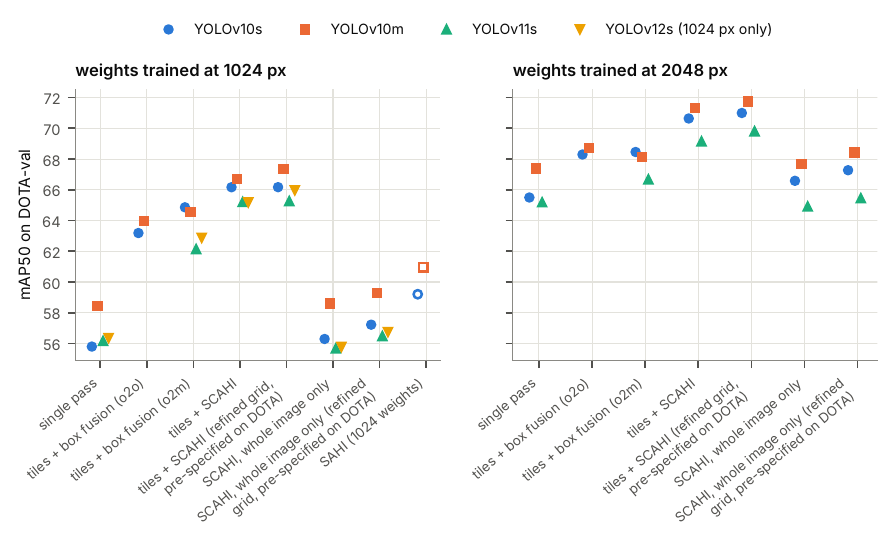}
  \caption{\textbf{All frozen configurations on DOTA-val} (mAP50, devkit, no cap), for YOLOv10s and YOLOv10m at 1024 and 2048\,px: single pass, tiles with box fusion (both heads), SCAHI with the setting from VisDrone and with the refined grid (pre-specified on DOTA), SCAHI on the single pass, and SAHI.}
  \label{fig:s-dota-rows}
\end{figure*}

\begin{figure*}[tp]
  \centering
  \includegraphics[width=0.8\textwidth]{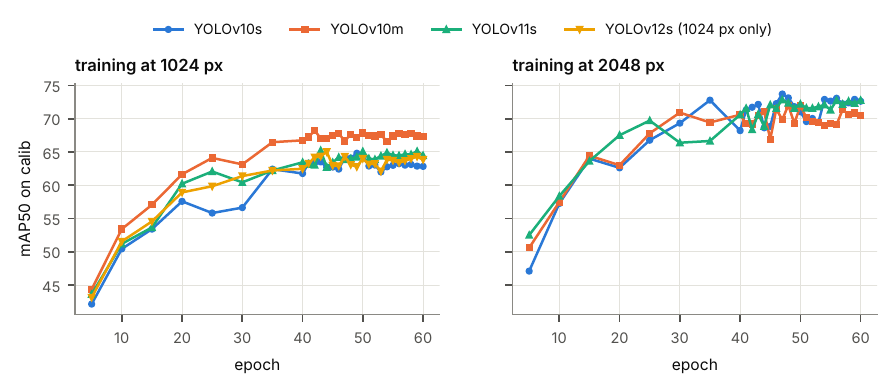}
  \caption{\textbf{Calibration mAP50 during training on DOTA}; the selected epoch is the maximum on calib.}
  \label{fig:s-dota-calib}
\end{figure*}

\textbf{Configurations of YOLOv10s/m.} \Cref{fig:s-dota-rows} shows all 32 frozen configurations. The decomposition of the gain differs
from VisDrone: on DOTA, adding tiles at native resolution contributes most (3.3 to 9.9\,mAP50), whereas fusing the single pass
alone contributes little. \Cref{fig:s-dota-calib} shows the calibration curves during training. \Cref{fig:s-dota-q} shows the pre-specified questions graphically.

\FloatBarrier
\section{SCAHI component ablation: details}
\label{sec:s-ablation}

\begin{figure*}[tp]
  \centering
  \includegraphics[width=0.8\textwidth]{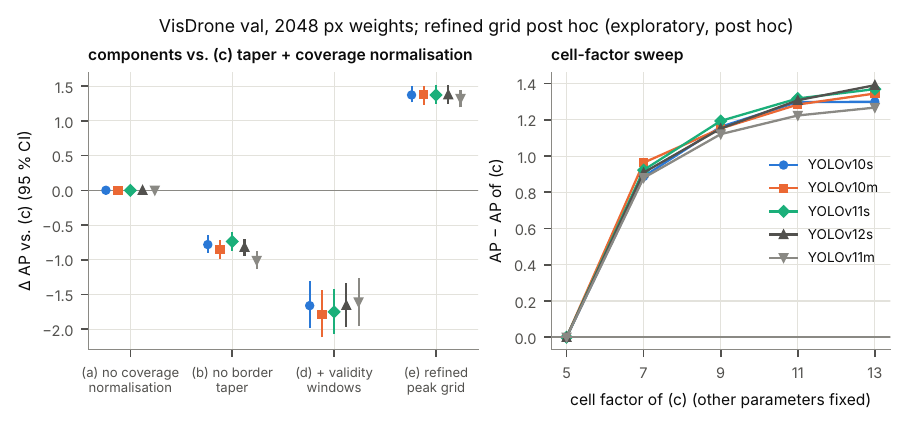}
  \caption{\textbf{Ablation on VisDrone val}: difference to the pre-specified SCAHI setting for each variant and detector, with paired bootstrap 95\,\% confidence intervals, and the sweep of the cell factor. All detectors of the original studies with 2048\,px weights; YOLOv12m could not be trained at 2048\,px.}
  \label{fig:s-abl-vis}
\end{figure*}

\begin{figure*}[tp]
  \centering
  \includegraphics[width=0.8\textwidth]{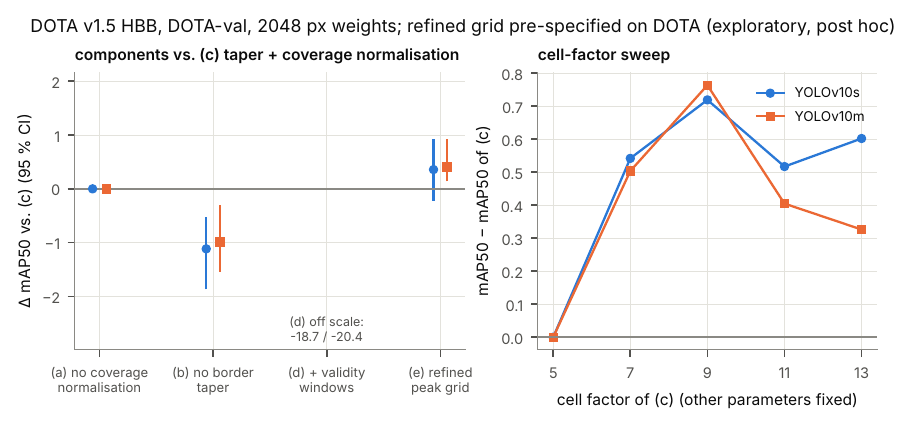}
  \caption{\textbf{Ablation on DOTA-val} (post hoc, after the single held-out evaluation). The validity windows are sized for VisDrone and cut off large DOTA objects; they are shown for completeness only.}
  \label{fig:s-abl-dota}
\end{figure*}

The ablation was run after the main results, from the existing detections, on VisDrone val and on DOTA-val (the latter after
its single held-out evaluation); it never enters a verdict. It covers the detectors available at that time; YOLO26, added
later, is not part of it. All variants use the same passes (single pass at 2048\,px plus
tiles) and the same parameter selection on calib. \Cref{fig:s-abl-vis,fig:s-abl-dota} show all variants per detector,
including a sweep of the cell factor of the peak grid. On VisDrone, accuracy rises almost monotonically with the cell factor;
on DOTA it peaks around a cell factor of 9, while the recall of crowded objects keeps rising.

\textbf{Why the normalisation has no effect.} We measured the coverage $C$ at every fused position. With the coverage-aware
tapers and two levels of passes that each cover the whole image, $C=2$ at every position on both datasets, so dividing by $C$
only rescales all scores. Without the taper, $C$ varies between 2 and 10 (mean about 3.4), and accuracy drops. That a pass
which covers a position but predicts nothing counts against an object follows from the summation over passes, not from the
division; the division would matter only for passes that cover part of the image, which our configurations do not contain.

\textbf{Border-weighted box fusion.} To separate the effect of the coverage-aware weights from that of merging dense outputs, we
gave box fusion the same weights (post hoc): every per-pass detection is scored with SCAHI's weight at its box centre and then
merged by the frozen NMS (variant a) or by Weighted Boxes Fusion (variant b), with the box-fusion parameters re-selected on
calib. \Cref{tab:s-wbf} compares these variants with plain box fusion and with SCAHI on the same passes and head. Weighted NMS
improves on plain box fusion, but SCAHI remains ahead on both datasets; weighted WBF falls below plain box fusion, because it
merges crowded neighbours.
\begin{table*}[tp]
\centering\scriptsize
\caption{\textbf{Border-weighted box fusion} (post hoc, same passes and head). VisDrone val: AP@[.5:.95] / AP50; DOTA-val:
mAP50 / AP@[.5:.95]. Recall at IoU 0.5 of crowded / isolated objects. Differences in the first metric with paired bootstrap
95\,\% CI, against box fusion and against SCAHI with the refined grid; only rows on the same head are compared. Variant (b) was
not run for FCOS.}
\label{tab:s-wbf}
\begin{tabular}{@{}lllrrrll@{}}
\toprule
\multicolumn{8}{@{}l}{\textit{VisDrone val}} \\
Detector & head & configuration & AP & AP50 & R50 crowded / isolated & $\Delta$ vs.\ box fusion & $\Delta$ vs.\ refined \\
\midrule
YOLOv10s & o2o & box fusion (frozen) & 38.85 & 64.73 & 79.8 / 91.0 &  &  \\
 &  & (a) weighted NMS & 39.57 & 64.75 & 86.2 / 91.5 & +0.72 [+0.52; +0.93] & -1.00 [-1.13; -0.84] \\
 &  & (b) weighted WBF & 37.56 & 62.34 & 87.1 / 93.0 & -1.29 [-1.57; -0.98] & -3.01 [-3.25; -2.74] \\
 &  & SCAHI, refined grid & 40.57 & 66.17 & 85.7 / 92.5 &  &  \\
\midrule
YOLOv10s & o2m & box fusion (post hoc, tuned like the frozen row) & 37.85 & 62.76 & 79.7 / 87.9 &  &  \\
 &  & (a) weighted NMS & 38.60 & 62.68 & 86.3 / 88.3 & +0.75 [+0.57; +0.95] & -2.12 [-2.29; -1.93] \\
 &  & (b) weighted WBF & 35.59 & 59.11 & 85.1 / 89.3 & -2.25 [-2.55; -1.90] & -5.12 [-5.42; -4.79] \\
 &  & SCAHI (pre-specified) & 39.34 & 64.19 & 63.4 / 89.0 &  &  \\
 &  & SCAHI, refined grid & 40.72 & 66.15 & 81.1 / 91.9 &  &  \\
\midrule
YOLOv11s & o2m & box fusion (frozen) & 37.22 & 62.18 & 79.1 / 87.6 &  &  \\
 &  & (a) weighted NMS & 37.72 & 62.52 & 77.3 / 87.3 & +0.50 [+0.30; +0.71] & -2.31 [-2.52; -2.14] \\
 &  & (b) weighted WBF & 35.12 & 58.64 & 84.4 / 89.1 & -2.10 [-2.42; -1.73] & -4.91 [-5.21; -4.57] \\
 &  & SCAHI (pre-specified) & 38.66 & 63.58 & 62.1 / 88.9 &  &  \\
 &  & SCAHI, refined grid & 40.03 & 65.30 & 81.6 / 91.6 &  &  \\
\midrule
\multicolumn{8}{@{}l}{\textit{DOTA-val}} \\
Detector & head & configuration & mAP50 & AP & R50 crowded / isolated & $\Delta$ vs.\ box fusion & $\Delta$ vs.\ refined \\
\midrule
YOLOv10s & o2m & box fusion (frozen) & 68.46 & 47.41 & 94.6 / 95.6 &  &  \\
 &  & (a) weighted NMS & 69.87 & 48.15 & 94.2 / 95.5 & +1.42 [+0.22; +1.72] & -1.12 [-2.27; -0.57] \\
 &  & (b) weighted WBF & 64.30 & 43.39 & 80.6 / 95.2 & -4.16 [-5.56; -2.71] & -6.70 [-8.04; -5.15] \\
 &  & SCAHI (pre-specified) & 70.64 & 49.15 & 66.8 / 96.9 &  &  \\
 &  & SCAHI, refined grid & 71.00 & 50.13 & 91.1 / 98.1 &  &  \\
\midrule
YOLOv11s & o2m & box fusion (frozen) & 66.68 & 45.93 & 92.5 / 95.5 &  &  \\
 &  & (a) weighted NMS & 68.41 & 47.02 & 89.4 / 95.2 & +1.73 [+0.55; +2.41] & -1.39 [-2.23; -0.72] \\
 &  & (b) weighted WBF & 63.32 & 42.80 & 81.4 / 96.3 & -3.36 [-4.47; -1.98] & -6.49 [-7.46; -5.05] \\
 &  & SCAHI (pre-specified) & 69.14 & 48.42 & 66.8 / 96.8 &  &  \\
 &  & SCAHI, refined grid & 69.81 & 49.36 & 94.5 / 98.0 &  &  \\
\midrule
FCOS & -- & box fusion (frozen) & 60.49 & 37.50 & 93.4 / 95.3 &  &  \\
 &  & (a) weighted NMS & 59.74 & 36.86 & 91.1 / 94.5 & -0.75 [-1.79; +0.14] & -0.85 [-2.24; +0.23] \\
 &  & SCAHI (pre-specified) & 60.59 & 37.55 & 62.6 / 93.9 &  &  \\
 &  & SCAHI, refined grid & 60.59 & 37.55 & 62.6 / 93.9 &  &  \\
\bottomrule
\end{tabular}
\end{table*}

\Cref{tab:s-params} lists the selected parameters of SCAHI for every configuration with tiles.
\begin{table*}[tp]
\centering\small
\caption{\textbf{Selected SCAHI parameters} for tiles with SCAHI (one-to-many head), per dataset, detector and size of the
single pass: cell factor $f$, bins per octave of the size axis, and the thresholds $\tau_\text{splat}$ and $\tau_\text{peak}$.
The pre-specified setting was tuned on calib over thresholds and bins with $f=5$; the refined grid also selects $f$ (post hoc
on VisDrone, pre-specified on DOTA). Predictions are clustered at IoU 0.5 in all settings.}
\label{tab:s-params}
\begin{tabular}{@{}lllrrrrrrrr@{}}
\toprule
 & & & \multicolumn{4}{c}{pre-specified setting} & \multicolumn{4}{c}{refined grid} \\
\cmidrule(lr){4-7}\cmidrule(lr){8-11}
Dataset & Detector & $N$ & $f$ & bins & $\tau_\text{splat}$ & $\tau_\text{peak}$ & $f$ & bins & $\tau_\text{splat}$ & $\tau_\text{peak}$ \\
\midrule
VisDrone & YOLOv10s & 2048 & 5 & 5 & 0.005 & 0.001 & 11 & 7 & 0.00125 & 0.001 \\
VisDrone & YOLOv10m & 1280 & 5 & 5 & 0.005 & 0.001 & 13 & 7 & 0.00125 & 0.001 \\
VisDrone & YOLOv10m & 2048 & 5 & 5 & 0.005 & 0.001 & 13 & 8 & 0.00031 & 0.001 \\
VisDrone & YOLOv11s & 1280 & 5 & 5 & 0.005 & 0.001 & 11 & 7 & 0.00125 & 0.001 \\
VisDrone & YOLOv11s & 2048 & 5 & 5 & 0.005 & 0.001 & 11 & 8 & 0.00031 & 0.001 \\
VisDrone & YOLOv11m & 1280 & 5 & 5 & 0.005 & 0.001 & 11 & 6 & 0.005 & 1e-05 \\
VisDrone & YOLOv11m & 2048 & 5 & 5 & 0.005 & 0.001 & 13 & 7 & 0.00125 & 0.001 \\
VisDrone & YOLOv12s & 1280 & 5 & 5 & 0.005 & 0.001 & 11 & 6 & 0.00031 & 0.001 \\
VisDrone & YOLOv12s & 2048 & 5 & 5 & 0.005 & 0.001 & 13 & 8 & 0.00125 & 0.001 \\
VisDrone & YOLOv12m & 1280 & 5 & 5 & 0.005 & 0.001 & 13 & 6 & 0.00031 & 0.001 \\
VisDrone & YOLO26s & 1280 & 5 & 5 & 0.005 & 0.001 & 13 & 7 & 0.00125 & 0.001 \\
VisDrone & YOLO26s & 2048 & 5 & 5 & 0.005 & 0.001 & 13 & 8 & 0.00125 & 0.001 \\
VisDrone & YOLO26m & 1280 & 5 & 5 & 0.005 & 0.001 & 13 & 8 & 0.00031 & 0.001 \\
VisDrone & YOLO26m & 2048 & 5 & 5 & 0.005 & 0.001 & 13 & 8 & 0.00125 & 0.001 \\
\midrule
DOTA & YOLOv10s & 1024 & 5 & 4 & 0.005 & 0.001 & 7 & 8 & 0.00031 & 0.001 \\
DOTA & YOLOv10s & 2048 & 5 & 4 & 0.005 & 0.001 & 7 & 7 & 0.00031 & 0.001 \\
DOTA & YOLOv10m & 1024 & 5 & 5 & 0.005 & 0.001 & 7 & 6 & 0.00125 & 1e-05 \\
DOTA & YOLOv10m & 2048 & 5 & 5 & 0.005 & 0.001 & 7 & 6 & 0.00125 & 0.0001 \\
\bottomrule
\end{tabular}
\end{table*}


\FloatBarrier
\section{Qualitative examples}
\label{sec:s-qual}

\begin{figure*}[tp]
  \centering
  \includegraphics[width=0.86\textwidth,height=0.86\textheight,keepaspectratio]{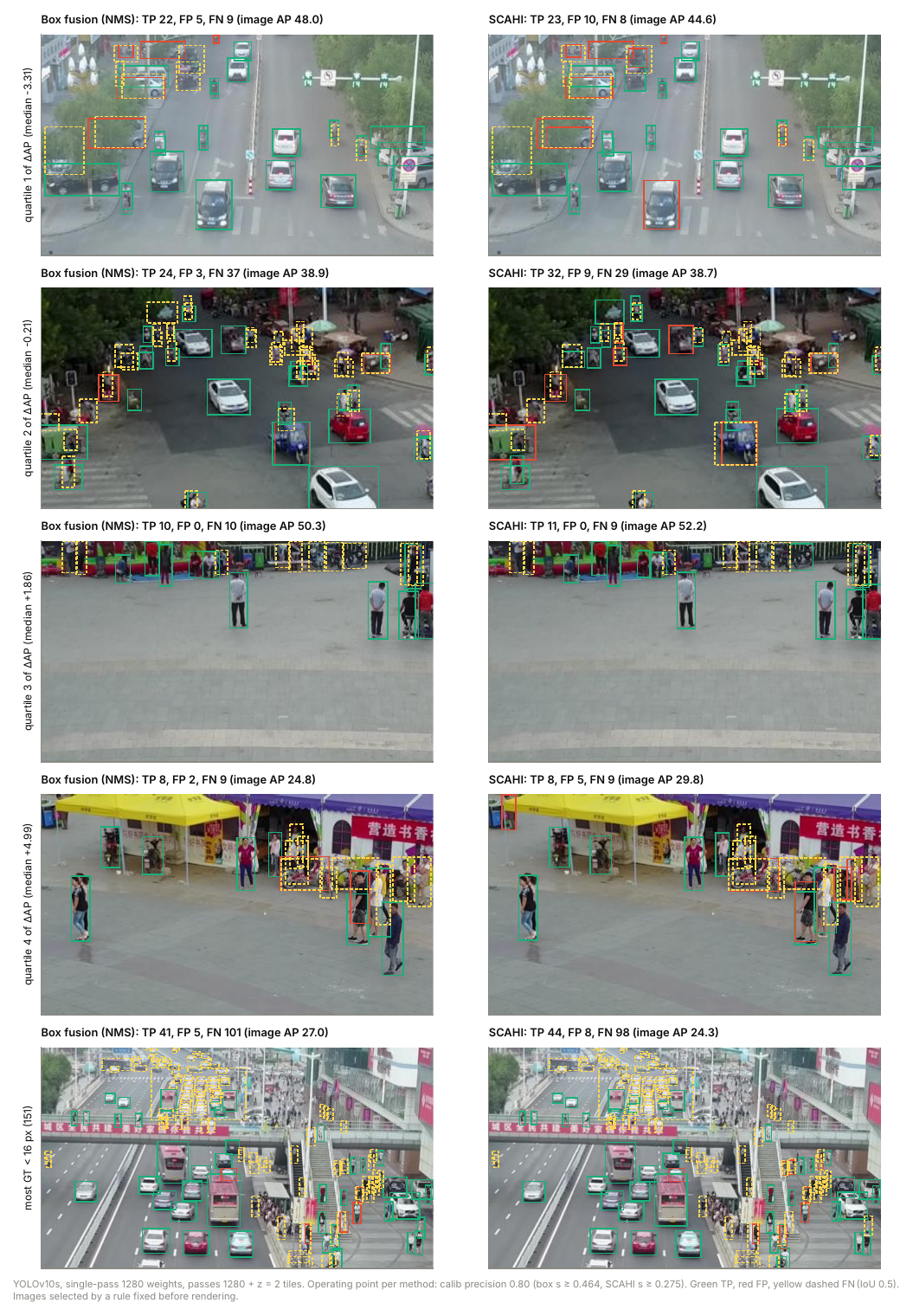}
  \caption{\textbf{Qualitative comparison} of box fusion (NMS, left) and SCAHI (pre-specified setting, right) on the passes of the 1280\,px pass with $z{=}2$ tiles (YOLOv10s, val). Green: correct detections; red: false positives; yellow dashed: missed objects (IoU 0.5). Rows: one image per quartile of the per-image AP difference, and the image with the most objects below 16\,px.}
  \label{fig:s-qual}
\end{figure*}

\Cref{fig:s-qual} compares box fusion and SCAHI on the same passes. To compare equal operating points, each method uses the
score threshold at which it reaches a precision of 0.80 on calib. The images were chosen by a rule fixed before rendering: one
image from each quartile of the per-image difference in AP between the two methods, and the image with the most objects below
16\,px. The examples show both directions: SCAHI recovers missed small objects in crowded scenes, but can also merge
neighbours, which is the failure mode addressed by the refined peak grid.

\begin{figure*}[tp]
  \centering
  \includegraphics[width=0.62\textwidth,height=0.87\textheight,keepaspectratio]{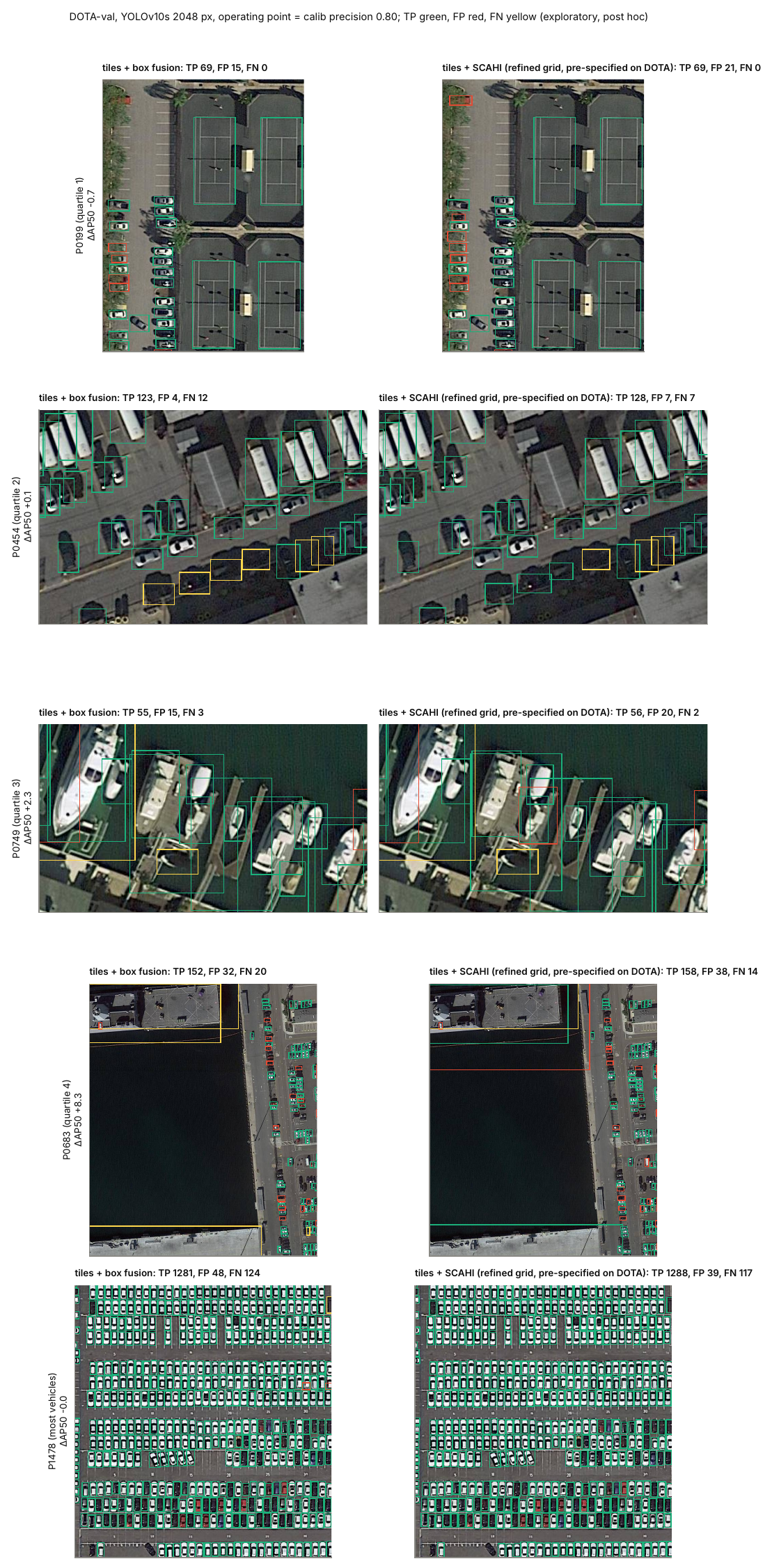}
  \caption{\textbf{Qualitative comparison on DOTA-val}: box fusion (left) and SCAHI with the refined grid (right), YOLOv10s, 2048\,px pass with tiles. Green: correct detections; red: false positives; yellow dashed: missed objects (IoU 0.5).}
  \label{fig:s-dota-qual}
\end{figure*}

\Cref{fig:s-dota-qual} shows the same comparison on DOTA, with operating points and image selection fixed in the same way
before rendering (one image per quartile of the per-image difference in mAP50 between SCAHI with the refined grid and box
fusion, YOLOv10s at 2048\,px, plus the image with the most vehicles). Dense parking lots show where box fusion loses crowded
vehicles.

\FloatBarrier
\section{Code and reproduction}
\label{sec:s-code}

\codelocation{} contains the complete code under the AGPL-3.0 licence\codelicencenote, together with the frozen
configurations of every reported row. It offers four entry points; the README gives the details.

\textbf{(a) Installation} with the exact package versions used for the paper, from the lock file:
\begin{footnotesize}
\begin{verbatim}
conda env create -f environment.yml
conda activate scahi
pip install -r requirements.lock
pip install -e .
\end{verbatim}
\end{footnotesize}

\textbf{(b) Regenerating the paper.} Every table of the paper and the data of every figure are regenerated from the stored
results, without datasets, weights or a GPU, and compared byte for byte with the expected outputs (about one minute on a CPU):
\begin{footnotesize}
\begin{verbatim}
python scripts/regenerate_paper.py
\end{verbatim}
\end{footnotesize}

\textbf{(c) SCAHI on one's own image.} The example runs the public COCO weights of YOLOv10s, which Ultralytics fetches by
model name, with the passes of the paper: one whole-image pass at a long side of 2048\,px and tiles at zoom $z=2$ (input
1024\,px, overlap 0.3). It fuses the dense one-to-many outputs with SCAHI, in the pre-specified or the refined setting, on the
CPU or the GPU, and optionally merges the same passes with box fusion for comparison:
\begin{footnotesize}
\begin{verbatim}
python examples/scahi_yolov10s.py img.jpg
python examples/scahi_yolov10s.py img.jpg \
    --setting refined --fusion-device gpu \
    --box-fusion --out dets.csv
\end{verbatim}
\end{footnotesize}

\textbf{(d) The full experiment pipeline} (training, caching of the dense outputs, tuning on calib, freezing, the single
evaluation) is documented in \texttt{REPRODUCE.md}, with the entry points of each study: the VisDrone main study and
extension, the DOTA study with the YOLO detectors and FCOS, SCAHI on SAHI's passes, the YOLO26 addendum, and the exploratory analyses. Trained
weights are not included because of their size.

\FloatBarrier
\section{Societal impact}
\label{sec:s-impact}

Object detection in aerial and UAV imagery supports beneficial applications such as search and rescue, disaster response,
traffic monitoring and infrastructure inspection. However, improving the detection and localisation of people and vehicles may
also facilitate intrusive surveillance and unauthorized monitoring. Although object detection does not establish individual
identities, its outputs could be combined with tracking or identification systems to infer sensitive information about
individuals and their movements. Our work does not identify individuals: it predicts class labels and boxes, and we used only
established public benchmarks (VisDrone, DOTA) as released by their creators, without collecting new data. Detection errors and
limited generalisation across environments may introduce additional risks in real-world use, and our results on public
benchmarks should not be read as evidence of suitability for high-stakes surveillance or automated decision-making. Responsible
deployment therefore requires appropriate privacy safeguards, data minimization, legal authorization and human oversight.
Finally, tiled inference multiplies the computation per image; where a single pass at a higher resolution suffices, as on
VisDrone in our experiments, it is also the more economical choice.

\FloatBarrier
\section{Deviations and disclosures}
\label{sec:s-dev}

All deviations are recorded in the decision log; none changed a pre-specified criterion after the corresponding results were
known.

\textbf{Dropped configurations.} YOLOv12m at 2048\,px did not fit into memory with a batch of at least two images, the
pre-specified minimum; it was dropped, and the affected questions count as not confirmed for YOLOv12m. The single pass at
3072\,px of the compute control did not fit with its standard batch and was dropped for the same reason.

\textbf{Recipe deviations.} YOLOv12s at 2048\,px and YOLOv12m at 1280\,px were trained with reduced batches (effective batch 64
via gradient accumulation), which affects the batch normalisation statistics. The YOLOv12m run was interrupted by an
out-of-memory error at epoch 66 and resumed from its checkpoint; its selected checkpoint (epoch 65) predates the interruption.

\textbf{Quality gates.} Two automated checks were corrected during the studies; a third, the single-pass gate of YOLO26s, was
not met as specified and is described in \cref{sec:s-y26}. A decoding-parity tolerance of $10^{-4}$\,px
was below float32 resolution for inputs above 1024\,px and became $\max(10^{-4}\,\text{px},\,1\,\text{ulp})$. A requirement of
identical AP to four decimals between our decoding and the reference failed for one detector because a single match flipped at
an IoU threshold; it became $|\Delta\text{AP}|\le0.01$ for all detectors. No verdict of YOLOv11m lies closer than 1.2\,AP to its threshold, so the choice of decoding cannot change it.

\textbf{Post-hoc analyses.} In the main study, one design comparison of the fusion was run on 60 val images before the
pre-specification and later repeated on calib. In the extension, the decomposition, the over-merging analysis and the refined
peak grid were defined after the val results were known; the refined grid was selected on calib and frozen before test-dev.
For the single-pass row of the refined grid, only the first search round was used, so its gain is likely underestimated.

\end{document}